\documentclass[pdflatex,sn-mathphys-num,iicol]{sn-jnl}

\usepackage{graphicx}%
\usepackage{multirow}%
\usepackage{amsmath,amssymb,amsfonts}%
\usepackage{amsthm}%
\usepackage{mathrsfs}%
\usepackage[title]{appendix}%
\usepackage[table, svgnames, dvipsnames]{xcolor}%
\usepackage{textcomp}%
\usepackage{manyfoot}%
\usepackage{booktabs}%
\usepackage{algorithm}%
\usepackage{algorithmicx}%
\usepackage{algpseudocode}%
\usepackage{listings}%

\usepackage{makecell}

\usepackage{tikz}
\usepackage{multirow}
\usepackage{multicol}
\definecolor{gold}{HTML}{FBF2D2}
\definecolor{silver}{HTML}{DDDDDD}
\definecolor{bronze}{HTML}{EED2B8}

\definecolor{goldD}{HTML}{D9AE13}
\definecolor{silverD}{HTML}{909090}
\definecolor{bronzeD}{HTML}{9A5F26}

\definecolor{catGreen}{HTML}{238763}
\definecolor{catBlue}{HTML}{1F70AE}

\definecolor{SuperSimpleNet_clr}{HTML}{a1c9f4}
\definecolor{SimpleNet_clr}{HTML}{377eb8}
\definecolor{BGAD_clr}{HTML}{4daf4a}
\definecolor{PRN_clr}{HTML}{984ea3}
\definecolor{SegDecNet_clr}{HTML}{ff7f00}
\definecolor{DRA_clr}{HTML}{ffff33}
\definecolor{AST_clr}{HTML}{66c2a5}
\definecolor{PatchCore_clr}{HTML}{fc8d62}
\definecolor{FastFlow_clr}{HTML}{8da0cb}
\definecolor{Draem_clr}{HTML}{e78ac3}
\definecolor{DSR_clr}{HTML}{a6d854}
\definecolor{EfficientAD_clr}{HTML}{ffd92f}

\definecolor{sup_mask_gen}{HTML}{82B366}

\definecolor{sn_part}{HTML}{5AC9A1}
\definecolor{ssn_part}{HTML}{0353A4}

\newcommand{\meanwithstd}[2]{\specialcell[c]{#1 \\[-2pt] {\footnotesize ($\pm$ #2)}}}
\newcommand{\specialcell}[2][c]{%
  \begin{tabular}[#1]{@{}c@{}}#2\end{tabular}}

\usepackage{xcolor}

\theoremstyle{thmstyleone}%
\theoremstyle{thmstyletwo}%

\theoremstyle{thmstylethree}%

\begin{document}

\title[No Label Left Behind: A Unified Surface Defect Detection Model for all Supervision Regimes]{No Label Left Behind: A Unified Surface Defect Detection Model for all Supervision Regimes}


\author*[1]{\fnm{Blaž} \sur{Rolih}}\email{blaz.rolih@fri.uni-lj.si}

\author[1]{\fnm{Matic} \sur{Fučka}}\email{matic.fucka@fri.uni-lj.si}

\author[1]{\fnm{Danijel} \sur{Skočaj}}\email{danijel.skocaj@fri.uni-lj.si}

\affil*[1]{\orgdiv{Faculty of Computer and Information Science}, \orgname{University of Ljubljana}, \orgaddress{\street{Večna Pot 113}, \city{Ljubljana}, \postcode{1000}, \country{Slovenia}}}

\abstract{Surface defect detection is a critical task across numerous industries, aimed at efficiently identifying and localising imperfections or irregularities on manufactured components. While numerous methods have been proposed, many fail to meet industrial demands for high performance, efficiency, and adaptability. Existing approaches are often constrained to specific supervision scenarios and struggle to adapt to the diverse data annotations encountered in real-world manufacturing processes, such as unsupervised, weakly supervised, mixed supervision, and fully supervised settings. To address these challenges, we propose \textit{SuperSimpleNet}, a highly efficient and adaptable discriminative model built on the foundation of SimpleNet. SuperSimpleNet incorporates a novel synthetic anomaly generation process, an enhanced classification head, and an improved learning procedure, enabling efficient training in all four supervision scenarios, making it the first model capable of fully leveraging all available data annotations. SuperSimpleNet sets a new standard for performance across all scenarios, as demonstrated by its results on four challenging benchmark datasets. Beyond accuracy, it is very fast, achieving an inference time below 10 ms. With its ability to unify diverse supervision paradigms while maintaining outstanding speed and reliability, SuperSimpleNet represents a promising step forward in addressing real-world manufacturing challenges and bridging the gap between academic research and industrial applications. Code: \textcolor{magenta}{https://github.com/blaz-r/SuperSimpleNet}.}




\maketitle

\section{Introduction}

Enhancing production efficiency to ensure high-quality products is a key priority in modern manufacturing. Detecting and addressing imperfections or irregularities on the surfaces of manufactured components plays a critical role in this effort. Historically, such detection has relied on manual inspection, a labour-intensive process prone to human error and inefficiencies~\citep{yang2025leis, tabernik2020segAD}. Automated systems, however, enable real-time monitoring, precise localisation of defects and anomalies, and significant improvements in product quality. Over the past decade, the field of computer vision has experienced remarkable progress, largely driven by advancements in deep learning paradigms. Unsurprisingly, these methods have also made their way into the traditionally conservative domain of machine vision, significantly advancing tasks such as visual inspection~\citep{mvtec, KSDD2, visa}. This progress exemplifies the power of collaboration between academic research and industrial applications, bridging the gap between cutting-edge technology and practical needs.

However, the core premises underlying academic research on the detection of surface irregularities do not always align perfectly with the practical requirements and scenarios encountered in manufacturing processes. The existing literature on this topic predominantly (with only a few exceptions) clusters into two distinct problem formulations. The first, \textit{surface defect detection (SDD)}~\citep{KSDD2,racki_sensum,tabernik2020segAD,bozic2021end2end}, assumes the availability of both normal and defective surface images during training, framing the task as a \textit{supervised} learning problem. The second, \textit{unsupervised anomaly detection (UAD)}~\citep{fuvcka_transfusion,zavrtanik_draem,zavrtanik2022dsr,batzner_efficientad}, is used in scenarios where only images of normal surfaces are available during training. Each approach has its merits. Supervised methods typically deliver superior performance~\citep{zavrtanik2022dsr, KSDD2}, making them the preferred choice whenever a sufficient number of labelled defect images is available. However, supervised models learn the decision boundary between normal and defective samples based on the training data, making them prone to missing novel or unseen defects that differ significantly from those in the training set~\citep{yao_bgad, zhang_prn}. In addition, labelling images is often a tedious task that disrupts the manufacturing process, and defective samples are frequently unavailable in sufficient quantities~\citep{KSDD2}. In contrast, UAD methods address these limitations by training exclusively on normal samples, allowing for the detection of previously unseen types of anomalies. This approach acknowledges the reality that normal samples are usually more abundant than defect samples in manufacturing environments~\citep{ding2022dra}. However, unsupervised methods generally perform worse than supervised approaches when comparable amounts of labelled data are available. Moreover, UAD methods are unable to leverage the information from any anomalous samples that may exist~\citep{ding2022dra}, which represents a significant limitation in many practical scenarios. In practice, at least some anomalous data typically exists at some stage, meaning this valuable information is completely disregarded in such cases.

We argue that both extremes—purely unsupervised or fully supervised methods-are not optimally suited for the complex task of detecting defects and anomalies on object surfaces in real-world applications. From an application-driven perspective, all available labels should be utilised. This necessitates the development of detection models capable of leveraging all possible supervision scenarios, combining the strengths of both approaches to better meet industrial demands.

\begin{figure}[t]
    \centering
    \includegraphics[width=1\linewidth]{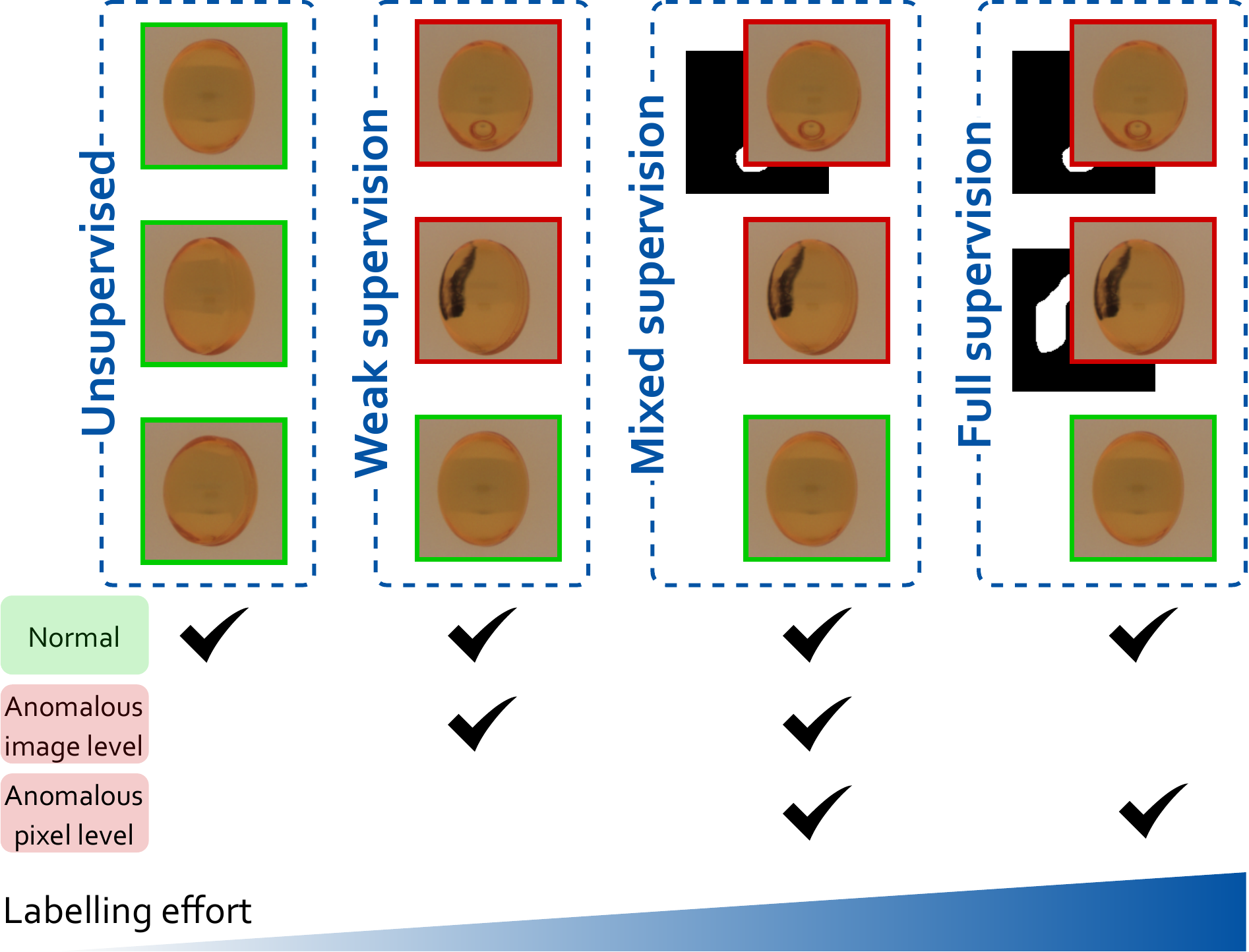}
    \caption{Different supervision scenarios within manufacturing processes are illustrated. Images with a \textcolor{catGreen}{green} border contain no anomalies, while images with a \textcolor{red}{red} border indicate the presence of an anomaly. For some images, the corresponding anomaly segmentation mask is also provided. The labelling effort required increases progressively from left to right. At present, only \textbf{SuperSimpleNet} supports training across all four scenarios.}  
    \label{fig:supervisions}
\end{figure}

\begin{table*}[t]
\centering
\resizebox{\textwidth}{!}{%
    \setlength{\tabcolsep}{2pt}
\begin{tabular}{l|cccc|ccccccccccccc}
\toprule
~ &US  &WS  &MS  &FS  &\textbf{\textcolor{red}{Ours}}  &SDNet  &TNet  &MMNet  &DRA  &EAD  &BGAD  &DSR  &SN  &FF  &PC  &DRÆM  &PRN \\\hline 
\textcolor{catGreen}{normal only} & \checkmark& ~& ~& ~& \checkmark& ~& ~& ~& ~& \checkmark& \checkmark& \checkmark& \checkmark& \checkmark& \checkmark& \checkmark& ~\\ 
\textcolor{red}{ano. image level}& ~& \checkmark& \checkmark& ~& \checkmark& \checkmark& \checkmark& \checkmark& \checkmark& ~& ~& ~& ~& ~& ~& ~& ~\\ 
\textcolor{red}{ano. pixel level}& ~& ~& \checkmark& \checkmark& \checkmark& \checkmark& \checkmark& \checkmark& ~& ~& \checkmark& \checkmark& ~& ~& ~& ~& \checkmark\\ 
\hline 
Speed ($<10ms$)& ~& ~& ~& ~& \checkmark& \checkmark& \checkmark& \checkmark& \checkmark& \checkmark& ~& ~& ~& ~& ~& ~& ~\\ 

 \bottomrule
\end{tabular}%
}
\caption{Four learning regimes utilising different levels of supervision: (i) \textcolor{catGreen}{normal images only}, \textcolor{catGreen}{normal} \textcolor{red}{and anomalous} images annotated with (ii) \textcolor{red}{image-level} and (iii) \textcolor{red}{pixel-level} labels. Twelve SOTA methods are listed, indicating which types of labels they are capable of processing. Additionally, the inference speed is reported, highlighting whether the requirement of $10ms$ is met. Notably, only our proposed method fulfils all these requirements. Models in the table are as follows: SDNet \citep{KSDD2}, TNet \citep{racki_sensum}, MMNet \citep{luo2023maminet}, DRA \citep{ding2022dra}, EAD \citep{batzner_efficientad}, BGAD \citep{yao_bgad}, DSR \citep{zavrtanik2022dsr}, SN \citep{liu_simplenet}, FF \citep{yu_fastflow}, PC \citep{roth_patchcore}, DRÆM \citep{zavrtanik_draem} and PRN \citep{zhang_prn}.}

\label{tab:requirements}
\end{table*}

To address these requirements, we introduce \textit{SuperSimpleNet}\footnote{Preliminary version presented in~\cite{rolih2024supersimplenet}.}, a novel discriminative model built upon the foundation established by SimpleNet~\citep{liu_simplenet}. 
SuperSimpleNet is designed to accommodate any available training data, as depicted in Fig.~\ref{fig:supervisions}, including\footnote{From now on, we will use the terms \textit{defect} and \textit{anomaly} interchangeably to emphasise the unification of the general problem of defect and anomaly detection.}: (i) \textit{normal images}, (ii) anomalous images annotated at the \textit{image level} (indicating whether an anomaly is present), and (iii) fully annotated images with \textit{pixel-level labels} that provide segmentation masks of anomaly regions. Based on these types of training data, we define four distinct training regimes: (i) \textit{unsupervised}, where only normal images are used; (ii) \textit{weakly supervised}, where anomalous images with image-level labels are also included; (iii) \textit{fully supervised}, where all images are pixel-level annotated; and (iv) \textit{mixed supervision}, where all images have image-level labels, but only a subset includes pixel-level segmentation masks.

The primary strength of SuperSimpleNet lies in its ability to effectively utilise all the aforementioned input types and operate seamlessly across the four training regimes. This flexibility ensures its applicability to a wide range of real-world industrial scenarios, bridging the gap between academic research and practical needs. To the best of our knowledge, it is the only method capable of handling all available input data, annotated at different levels of supervision, while achieving good results across all supervision scenarios. Additionally, it is highly efficient, addressing a key requirement frequently encountered in industrial applications. This is illustrated in Table~\ref{tab:requirements}, which highlights the main properties of various methods for defect and anomaly detection, along with their respective capabilities.

To achieve these goals, we significantly enhanced SimpleNet in terms of efficiency, effectiveness, and flexibility. In addition to several technical refinements that improved the original method, the key scientific contributions of our work are as follows:
\begin{itemize}
\item \textit{Improved Synthetic Anomaly Generation:} We propose a novel synthetic anomaly generation process that substantially enhances training in unsupervised scenarios and facilitates effective training in situations with limited or no pixel-level annotations. This innovation also strengthens segmentation capabilities, enabling robust and reliable predictions in cases where traditional methods often struggle.

\item \textit{Effective Classification Head:} We design a simple yet highly effective classification head that captures the global context of images and enables efficiently utilisation of image-level annotations. This addition enhances the model's ability to process diverse types of labelled data, improving its adaptability.

\item \textit{A Unified Model for All Supervision Paradigms:} By integrating techniques and insights from different levels of supervision into a single architecture, we present a unified approach that facilitates knowledge transfer across regimes. This significantly improves the performance of defect detection and anomaly segmentation, offering a versatile and scalable solution tailored to real-world industrial challenges. 

\end{itemize}

We conducted extensive experiments across four challenging datasets, demonstrating the versatility and performance of SuperSimpleNet. In the fully supervised setting, SuperSimpleNet achieved state-of-the-art results on two standard real-world defect detection datasets, SensumSODF~\citep{racki_sensum} and KSDD2~\citep{KSDD2}, with AUROC scores of 98.0\% and detection AP of 97.8\%, respectively. In the weakly supervised setting, SuperSimpleNet achieved an AUROC of 97.4\% on SensumSODF, outperforming previous fully supervised SOTA and a detection AP of 97.2\% on KSDD2. Similarly, it consistently outperforms existing methods on both SensumSODF and KSDD2 in the mixed supervised setting. For unsupervised anomaly detection, we tested SuperSimpleNet on two standard benchmarks, MVTec AD~\citep{mvtec} and VisA~\citep{visa}, where it achieved AUROC scores of 98.3\% and 93.6\%, respectively, again matching state-of-the-art performance. In addition to its superior accuracy, SuperSimpleNet demonstrated exceptional efficiency, with an inference time of 9.5~ms and a throughput of 262 images per second, surpassing most contemporary models in speed. Its efficiency, adaptability and outstanding results across all four settings—fully supervised, weakly supervised, mixed supervised, and unsupervised—highlight its effectiveness and make it a robust solution for a wide range of real-world anomaly detection scenarios.

\section{Related Work}

\noindent\textbf{Unsupervised methods}

\noindent Unsupervised anomaly detection has become a prominent research direction in recent years and has received a lot of attention. Those approaches are particularly useful in scenarios such as setting up new production lines or other situations where normal samples are abundant, but anomalous samples are rare. Among unsupervised methods, \textit{reconstruction-based approaches} initially gained popularity. These methods typically train an autoencoder-like (AE-like) network~\citep{zavrtanik_riad, ae-ssim}, assuming the reconstructed image will differ from the input image on the anomalous regions. Models from this paradigm are not limited to AE networks but also use other generative networks, such as GANs~\citep{akcay_ganomaly}, transformers~\citep{intra} or diffusion models~\citep{wyatt_noddpm}. The assumption about the successful reconstruction of anomalous regions does not always hold, leading to a decrease in overall performance.

Another successful approach in the recent past involves methods that leverage \textit{features} extracted from \textit{pretrained networks}, such as ResNet~\citep{he_resnet}. These extracted features are then utilised in various ways, including through memory-bank~\citep{roth_patchcore}, distillation techniques~\citep{reverse_dist, rudolph_ast}, student-teacher architectures~\citep{zhang_destseg} or normalizing flows~\citep{rudolph_ast, yu_fastflow}.

The last distinctive group consists of \textit{discriminative methods} that are trained using synthetic anomalies. The techniques of generating anomalies have evolved significantly over the past few years, progressing from simple cut-and-paste methods~\citep{li2021cutpaste} to more advanced approaches like Perlin noise~\citep{zavrtanik_draem} and diffusion-based anomaly generation~\citep{Zhang2024realNet}. These anomalies can be generated either directly on input images~\citep{zavrtanik_draem, zhang_destseg, fuvcka_transfusion} or, as more recently shown, at the feature-level~\citep{zavrtanik2022dsr, liu_simplenet}. 

While unsupervised methods achieve great detection results, they often struggle to simultaneously achieve fast operation times, prompting recent research to focus on more efficient solutions~\citep{batzner_efficientad, rolih2024tiledEns, ficsne2024fast}.

\noindent\textbf{Fully Supervised Methods}

\noindent Although anomalous samples are initially rare or unavailable, they gradually accumulate over time in industrial settings. Usually, these samples do not represent the entire distribution of potential anomalies, but research has shown that their use can improve detection performance~\citep{zavrtanik2022dsr}. Most unsupervised methods are not designed to leverage anomalous data, leading to the development of fully supervised methods, such as SegDecNet~\citep{KSDD2}, TriNet~\citep{racki_sensum}, and MaMiNet~\citep{luo2023maminet} in industrial applications to maximise the detection performance.

Recent approaches, such as BGAD~\citep{yao_bgad} and PRN~\citep{zhang_prn}, attempt to enhance anomaly detection by generating additional synthetic anomalies, further diversifying the anomalies the model encounters. However, their complex architecture results in longer inference times, limiting their suitability for industrial environments. 

Moreover, most fully supervised methods still require numerous fully-labelled samples to outperform the most recent unsupervised methods. They also lack support for strictly unsupervised learning, making them impractical in certain real-world scenarios, such as setting up a new production line.

\noindent \textbf{Weakly Supervised Methods}

\noindent In contrast to fully supervised methods, weakly supervised approaches require only image-level labels, demanding much less labelling effort than pixel-level annotations. The first significant attempt at weakly supervised anomaly detection was made by DeepSAD~\citep{ruff_deepSAD}, a one-class classification method that utilises image-level labels during training. A similar strategy was recently employed by SegAD~\citep{baitieva2024supervised}, which trains a weakly supervised random forest classifier using the predictions of deep unsupervised methods.

A key paradigm in weakly supervised anomaly localisation involves using class activation maps (CAM)~\citep{Zhou2016CAM}, which allows neural networks to localise anomalies without pixel-level supervision. Due to this property, several methods~\citep{zhang2021CADN, pang2021devnet, wu2022camBasedWeakly, jiang2024weakly} have adopted CAM for weakly supervised anomaly detection.

Despite the absence of pixel-level labels, some weakly supervised methods indirectly train for localization by leveraging techniques such as entropy minimization~\citep{xu2020cnnWeakly}, training exclusively on anomaly-free images~\citep{KSDD2, luo2023maminet}, or generating pseudo-labels from unsupervised models~\citep{ravcki2024unsupSup}.

\noindent \textbf{Mixed Supervision Methods}

\noindent Mixed supervision in anomaly detection has received limited attention in research. One of the pioneering approaches in this area is SegDecNet~\citep{KSDD2, tabernik2020segAD}, which employs a dual-head architecture — one head for detection and the other for segmentation. This foundational model has been extended in later work~\citep{luo2023maminet} by incorporating attention mechanisms to enhance performance. However, these models still don't support unsupervised training.

Unlike previous approaches, SuperSimpleNet can perform aptly in all four supervision scenarios. It achieves great detection performance even in unsupervised and weakly supervised settings. 
Earlier methods either lack support for all four training regimes or require pixel-level annotations to enable segmentation.

\section{SuperSimpleNet}

\begin{figure*}[t!]
    \centering
    \includegraphics[width=1\linewidth]{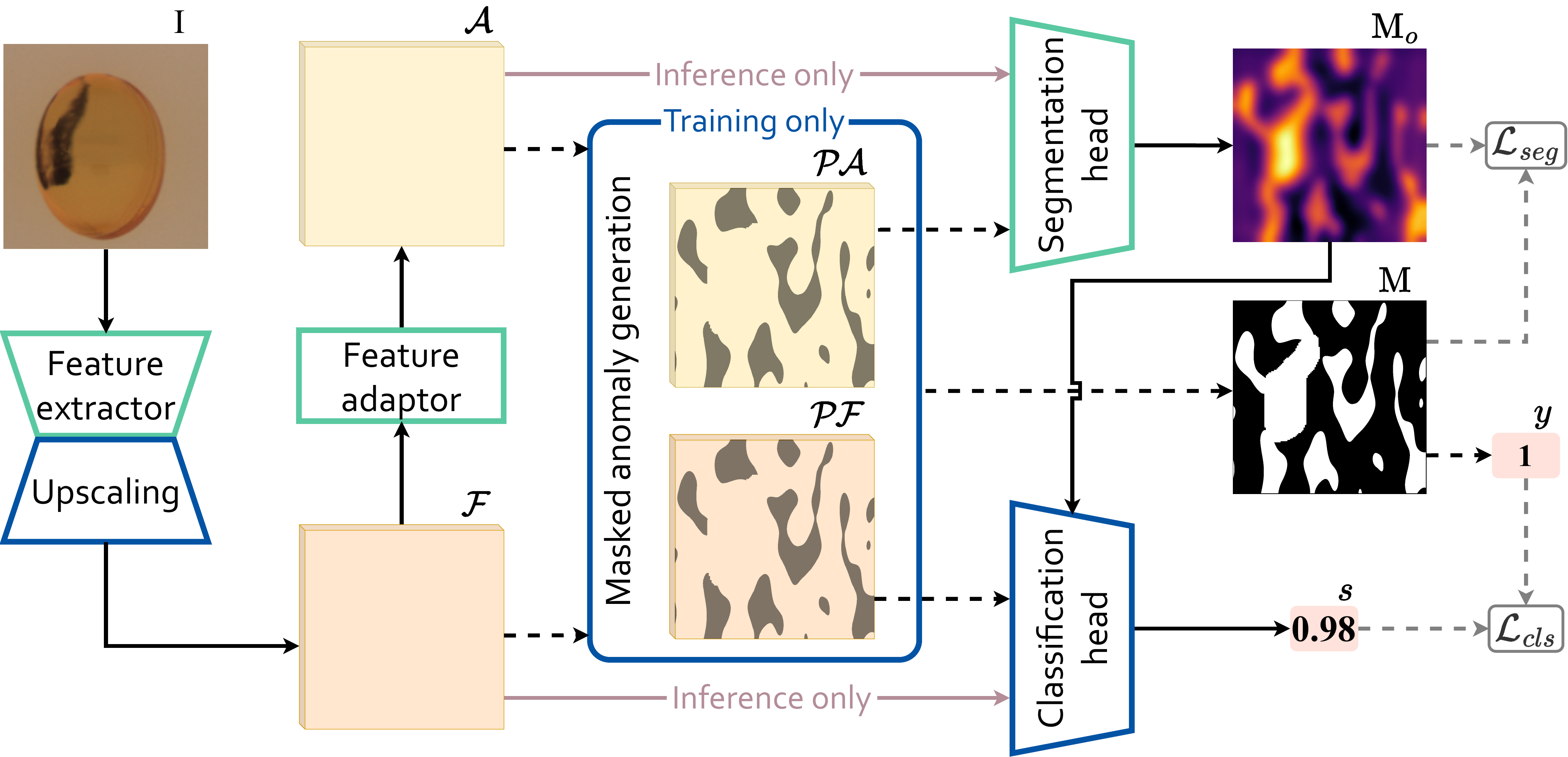}
    \caption{
    SuperSimpleNet's architecture. Features are first extracted, upscaled, and, in the case of the segmentation branch, also adapted. During training, synthetic anomalies are generated in the latent space and are limited to regions defined by the binarised Perlin mask. The segmentation head predicts an anomaly mask $\mathrm{M}_o$, based on the perturbed feature map $\mathcal{PA}$, which is then used in combination with the perturbed feature map $\mathcal{PF}$ by the classification head to produce the anomaly score $s$. The anomaly score $s$, and the predicted map $\mathrm{M}_o$ are supervised by the anomaly mask $\mathrm{M}$, and the ground truth anomaly score $y$, where $y$ is set to 1 if the image contains an anomaly (synthetic or real) and to 0 otherwise. During inference, $\mathrm{M}_o$ and $s$ are produced directly, skipping the anomaly generation phase. The remaining parts of original SimpleNet are also shown in the image with \textcolor{sn_part}{green} colour.}
    \label{fig:arch}
\end{figure*}

SuperSimpleNet is depicted in Figure~\ref{fig:arch}. First, features are extracted from the input image using a pretrained convolutional network. These are then upscaled and pooled to capture the neighbouring context (Section~\ref{sec:feat_extr}). Since anomaly segmentation and anomaly detection are distinct tasks, we use a simple adaptor to adapt the features for segmentation. Synthetic anomalies are then injected into both the segmentation and detection features. The key advancement of SuperSimpleNet is its novel synthetic anomaly generation mechanism, which enables strong performance across all supervision regimes. In the unsupervised setting, SuperSimpleNet exclusively depends on generated synthetic anomalies and their labels. In other regimes, synthetic anomalies complement the real ones, enriching the training signal and improving overall performance. The new strategy creates more realistic anomaly regions at the feature level by leveraging a binarised Perlin noise mask (Section~\ref{sec:ano_gen}). The processed features are then used in the segmentation and detection module (Section~\ref{sec:seg_dec}), whose dual-branch design enables learning with mixed supervision (Section~\ref{sec:loss}). All training mechanisms are discarded during inference, allowing the network to swiftly predict an anomaly map and score (Section~\ref{sec:inference}).

The following sections describe all the mentioned modules in detail, including training specifics with mixed supervision, to provide a clear understanding of SuperSimpleNet.

\subsection{Feature extractor}
\label{sec:feat_extr}

Following SimpleNet~\citep{liu_simplenet}, the input image $I$ is first passed through the pretrained feature extractor $\Phi$ (in our case, WideResnet50~\citep{zagoruyko_wideresnet} pretrained on ImageNet~\citep{deng_imagenet}) to extract features $f_l$ from a subset of layers $L$ where $l \in L$ and in our case $L = \{2,3\}$. Due to the architectural design of ResNet-like~\citep{he_resnet} networks, these features have relatively low spatial resolution. This limits the model's ability to detect smaller anomalies and reduces localisation precision. To address this limitation, an upscaling layer is introduced before feature concatenation, that is:
\begin{equation}
    F_l = \text{Upsample}(f_l, (2 H_0, 2 W_0)) \enspace ,
\end{equation}
where $(H_0, W_0)$ is the size of the largest extracted feature map, $l \in L$ and bilinear interpolation is used. This effectively doubles the feature resolution. In our case, this means that the 3rd layer's features are enlarged by a factor of 4 and the 2nd layer's features are enlarged by a factor of 2. This ensures that all layers match in spatial resolution and can be concatenated to produce $\hat{\mathcal{F}}$:
\begin{equation}
    \hat{\mathcal{F}} = \text{Concatenate}(\{F_l \mid l \in L \}) \enspace .
\end{equation}
 
In line with SimpleNet~\citep{liu_simplenet}, the neighbouring context for each feature location is captured using local average pooling with a $3\times3$ kernel, that is:
\begin{equation}
    \mathcal{F} = \text{AvgKernel}(\hat{\mathcal{F}}, \text{kernel size = 3}, \text{stride} = 1) \enspace .
\end{equation}
This all results in an upscaled feature map, where each location encodes contextual information. 

Although features from networks trained on natural images transfer well to the anomaly detection task~\citep{heckler_feature_importance}, a simple linear layer is used to adapt these features to a task-specific latent space. Only the features used as the input for the segmentation head are adapted ($\mathcal{A}$ in Figure~\ref{fig:arch}). We hypothesise that this is more effective because it allows segmentation and classification heads to operate in distinct latent spaces, leading to better specialisation for each task.

\subsection{Latent-space masked anomaly generation}
\label{sec:ano_gen}

In the \textit{supervised setting}, while training solely with the real anomalous data is possible, it may not fully capture the distribution of potential defects. 
To address this, we propose a novel synthetic anomaly generation technique inside the latent space of a pretrained feature extractor capable of generating highly randomised anomalies. As a result, all training regimes besides the unsupervised regime utilise a mix of real and synthetic anomalies. Furthermore, the new synthetic anomaly generation enables the model to be trained in all four supervision regimes while simultaneously being able to learn anomaly segmentation despite not having any available labelled data. This enables SuperSimpleNet to bridge the performance gap between fully and weakly supervised settings, where many previous methods struggle.

\begin{figure}[t]
    \centering
    \includegraphics[width=1\linewidth]{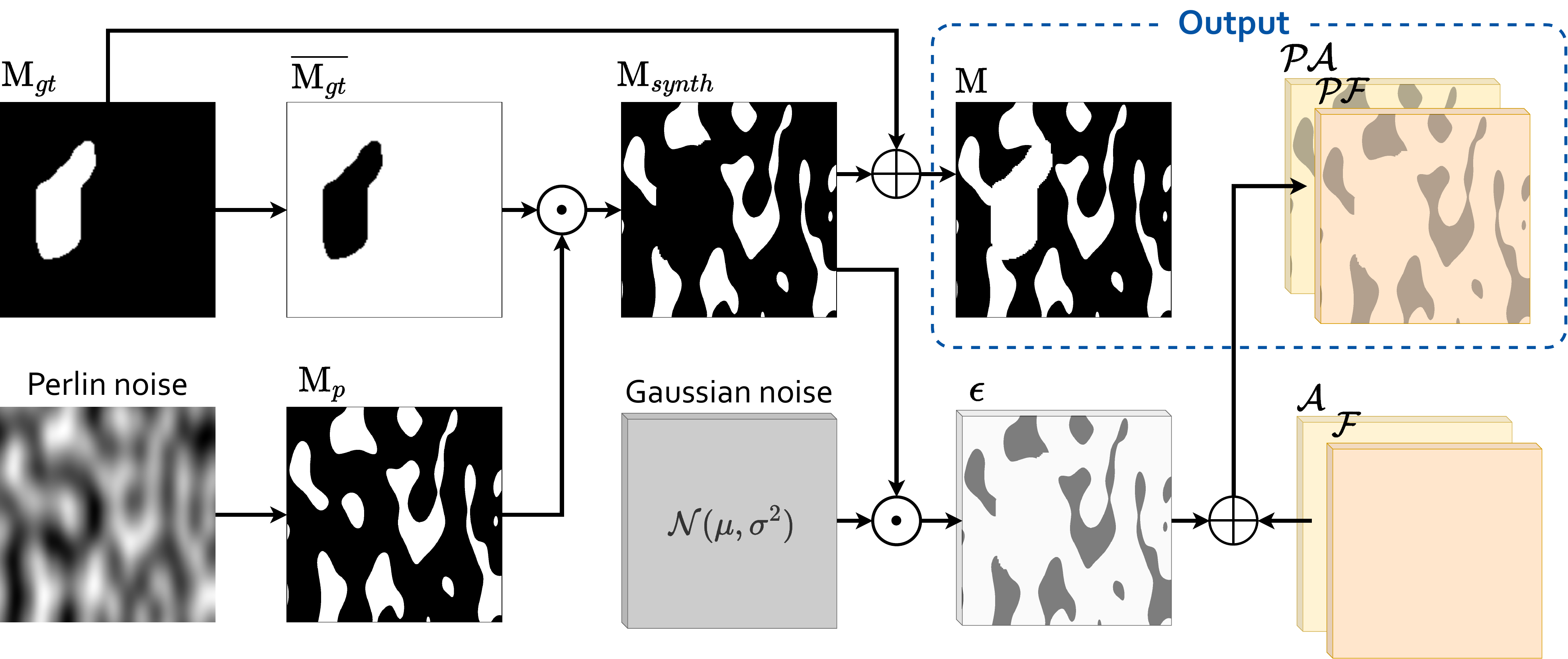}
    \caption{Synthetic anomaly generation. Synthetic anomaly masks $\mathrm{M}_{synth}$ are generated by removing actual anomalous regions (captured by ground truth mask $\mathrm{M}_{gt}$) from Perlin anomaly mask $\mathrm{M}_p$ (obtained by thresholding Perlin Noise~\citep{perlin1985image}). 
    $\mathrm{M}_{synth}$ is then used to limit the Gaussian noise only to specific regions, producing final noise $\epsilon$, which is later added to the features to create synthetic anomalies.
    The final anomaly mask $\mathrm{M}$ is constructed from $\mathrm{M}_{synth}$ and $\mathrm{M}_{gt}$ indicates regions with synthetic and actual anomalies. Since $\mathrm{M}_{gt}$ is empty in the case of weakly supervised and unsupervised learning, $\mathrm{M}_p$ directly becomes $\mathrm{M}_{synth}$ and the final mask $\mathrm{M}$.}
    \label{fig:noise}
\end{figure}

The new synthetic anomaly generation strategy is illustrated in Figure~\ref{fig:noise}. The core idea is to generate synthetic anomalies using Gaussian noise, which is applied only to regions defined by the synthetic anomaly mask $\mathrm{M}_{synth}$. To obtain this mask, Perlin noise~\citep{perlin1985image} is generated (as in~\cite{zavrtanik_draem, fuvcka_transfusion, zhang_destseg}) and then thresholded, resulting in a Perlin anomaly mask $\mathrm{M}_p$. To get the final synthetic anomaly mask $\mathrm{M}_{synth}$, the regions corresponding to actual anomalies, delineated by $\mathrm{M}_{gt}$, are removed from the Perlin anomaly mask $\mathrm{M}_p$, that is: 
\begin{equation}
    \mathrm{M}_{synth} = \mathrm{M}_{p} \cdot \overline{\mathrm{M}_{gt}} \enspace .
\end{equation}
In the unsupervised or the weakly supervised setting, $\mathrm{M}_{gt}$ is always empty, so $\mathrm{M}_p$ directly serves as $\mathrm{M}_{synth}$.

Gaussian noise, sampled from the distribution $\mathcal{N}(\mu, \sigma^2)$, is restricted to regions defined by the synthetic anomaly mask $\mathrm{M}_{synth}$ to produce $\epsilon$ in Figure~\ref{fig:noise}. This region-limited noise is later added to both $\mathcal{F}$ and $\mathcal{A}$, resulting in the perturbed feature map $\mathcal{PF}$ and perturbed adapted feature map $\mathcal{PA}$, respectively. This approach ensures that SuperSimpleNet generates more realistic, spatially coherent, and highly randomised anomalous regions. The randomness of this process also prevents overfitting to unrealistic patterns that may not represent unseen data.

We retain the feature duplication mechanism from SimpleNet but apply noise to both the original and the duplicated features following the previously described process. This stabilises the training as the model is exposed to a wider variety of anomalies in each batch.

\subsection{Segmentation-detection module}
\label{sec:seg_dec}

The architecture of SimpleNet~\citep{liu_simplenet} is extended with a new \textit{classification head}, $D_{cls}$, while retaining the original \textit{segmentation head}, $D_{seg}$. Consistent with the overall simplicity of the architecture, the classification head consists of a single $5\times5$ convolutional block, followed by pooling layers and a final fully connected layer. Despite its simplicity, this design offers strong discriminative power, allowing the model to better capture the global context. This leads to a reduction in false positives and improved detection of small and in-distribution defects. 

The simplicity of the design is important in achieving good performance across both unsupervised and supervised settings, providing sufficient discriminative power while minimising the risk of overfitting. Moreover, the dual-branch design is well-suited for mixed supervision, as the network can be trained efficiently just with image-level labels.

\begin{figure}
    \centering
    \includegraphics[width=0.9\linewidth]{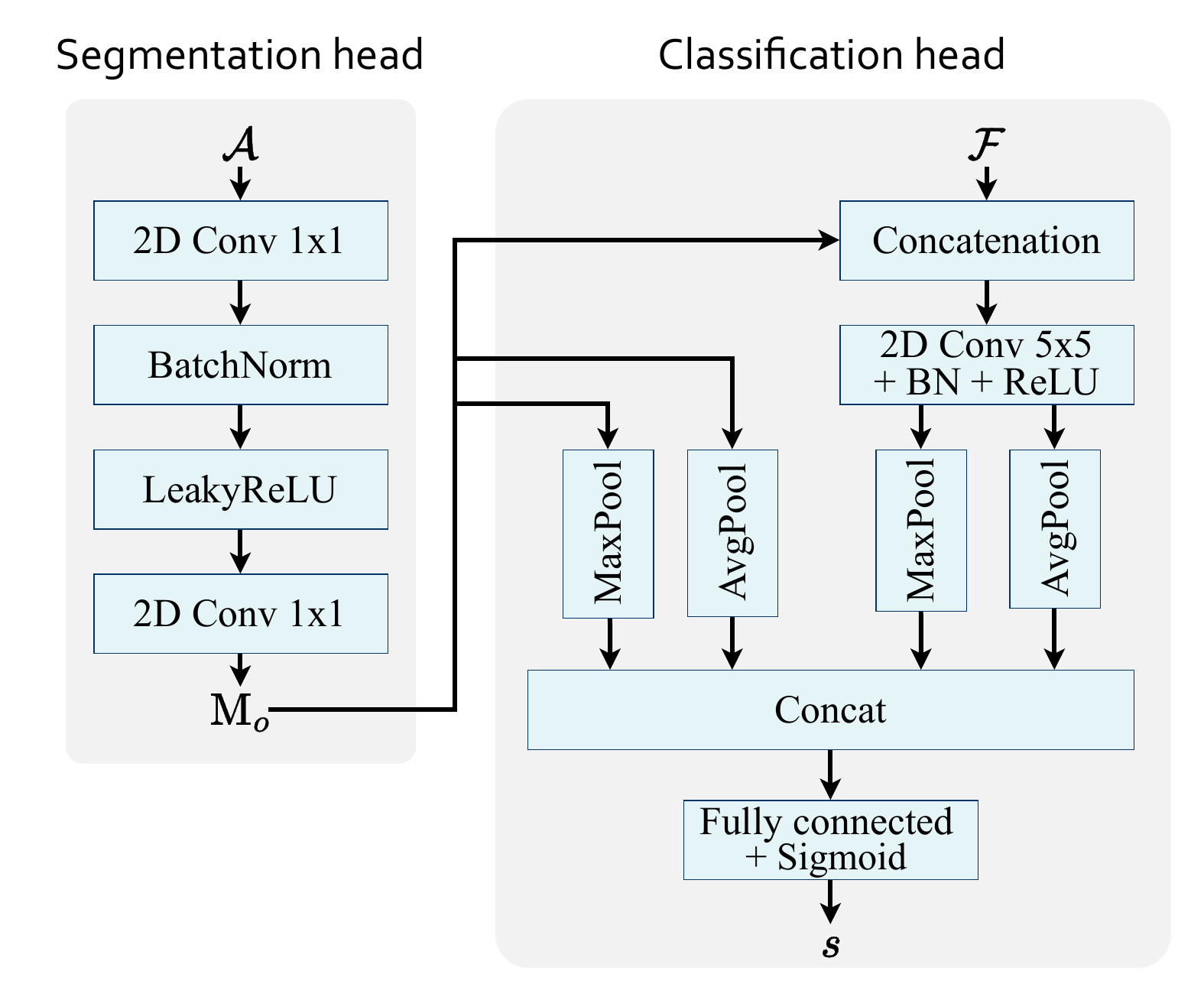}
    \caption{Detailed architecture of the segmentation-detection module. The design preserves the segmentation head from SimpleNet while introducing a new classification head with a wider kernel. This design allows for better contextual understanding, improving anomaly detection capabilities. }
    \label{fig:segdec}
\end{figure}

As shown in Figure~\ref{fig:segdec}, the segmentation head first generates an anomaly map $\mathrm{M}_o$. This map is then concatenated with the feature map $\mathcal{F}$ (or noise-augmented feature map $\mathcal{PF}$ during training) and serves as input to the classification head's convolutional block. The output from the convolutional block and the anomaly map are pooled, concatenated, and passed through the final fully connected layer to produce an image-level anomaly score $s$.

\subsection{Training with mixed supervision}
\label{sec:loss}

The truncated $\mathcal{L}_1$ loss is used for the segmentation head:
\begin{equation}
    l_{i,j} = 
    \begin{cases}
        max(0, th - D_{seg}(P_{i, j})) \text{; if } \mathrm{M[i,j]} = 1\\
        max(0, th + D_{seg}(P_{i, j})) \text{; otherwise}
    \end{cases}
    \enspace ,
    \label{eq:l1_parts}
\end{equation}
where $th$ is the truncation term preventing overfitting (in our case, 0.5), $D_{seg}$ is the segmentation head and $P_{i, j}$ is the value of the predicted anomaly mask at the position $(i,j)$.
The total truncated $\mathcal{L}_1$ loss, denoted by $\mathcal{L}_{1t}$, is computed as the mean of terms $l_{i,j}$ across all elements within the predicted anomaly mask. This loss encourages the model to learn a soft decision boundary between anomalous and non-anomalous regions. Due to the soft decision boundary, the model does not overfit to the data, resulting in better generalisation.
We additionally use focal loss~\citep{lin_focal} due to its good performance in cases of unbalanced data, resulting in the formulation of the final segmentation loss, denoted as $\mathcal{L}_{seg}$:
\begin{equation}
    \mathcal{L}_{seg} = \mathcal{L}_{1t} + \mathcal{L}_{foc} \enspace .
    \label{eq:seg_loss}
\end{equation}
\noindent Focal loss~\citep{lin_focal} is also used as the classification loss $\mathcal{L}_{cls}$:
\begin{equation}
    \mathcal{L}_{cls} = \mathcal{L}_{foc} \enspace .
    \label{eq:cls_loss}
\end{equation}
The final loss combines the segmentation and classification losses, with control term $\gamma$ added:
\begin{equation}
    \mathcal{L} = \gamma \cdot \mathcal{L}_{seg} + \mathcal{L}_{cls} \enspace .
    \label{eq:total_loss}
\end{equation}
The $\gamma$ term enables learning with mixed supervision and is determined based on the image label, following these rules: 
\begin{equation}
    \gamma = 
    \begin{cases}
        1 \text{; if image is normal;}\\
        1 \text{; if image is ano. \& fully labelled;}\\
        0 \text{; if image is ano. \& weakly labelled;}\\
    \end{cases}
\end{equation}
This allows the segmentation head to be trained on all images except anomalous images without pixel-level labels, while the classification head is always trained. Thanks to the new synthetic anomaly generation strategy, the model can successfully train segmentation even in the total absence of pixel-level labels.

Following the approach of SegDecNet~\citep{KSDD2}, we apply segmentation loss weighting to emphasise pixels in the centre of anomalous regions while reducing the focus on the more uncertain border pixels. The weights are produced with the distance transform applied to the ground truth mask, and are then simply multiplied by the loss~\citep{KSDD2}. This approach efficiently addresses labelling uncertainty in the boundary areas.

The target for the segmentation loss is the anomaly mask $\mathrm{M}$, which marks areas containing both synthetic and real anomalies. The target anomaly label $y$ for classification loss is set to 1 if the image contains an anomaly (either synthetic or real) and 0 otherwise.

\subsection{Inference}
\label{sec:inference}

In the inference stage, the anomaly generation mechanism is removed, and the model directly predicts the anomaly map using the segmentation head and anomaly score using the classification head. The anomaly map is upscaled to match the size of the input image and refined using a Gaussian filter with $\sigma=4$, yielding the final anomaly map.

\section{Experiments}

This section outlines the evaluation setup, including details on the data\-sets, metrics and implementation, followed by a presentation and discussion of the main results and ablation study results.

\subsection{Datasets}
\label{subsec:datasets}

The performance of SuperSimpleNet across \textit{fully supervised}, \textit{mixed supervised} and \textit{weakly supervised} settings is evaluated using two real-world, well-annotated datasets: Sensum Solid Oral Dosage Forms (SensumSODF)~\citep{racki_sensum} and Kolektor Surface-Defect Dataset 2 (KSDD2)~\citep{KSDD2}. 
\textbf{SensumSODF} consists of two categories representing different types of solid oral dosage forms: softgels and capsules. Each category includes normal and precisely annotated anomalous images featuring defects of various complexities and sizes. Since the dataset lacks a predefined train-test split, we follow the original protocol and perform 3-fold cross-validation~\citep{racki_sensum}. Figure~\ref{fig:sup_qual} shows an example from both categories.
\textbf{KSDD2} contains images of production items captured using a visual inspection system, with both train and test sets containing normal and annotated anomalous images. The defects present are often visually similar to good regions, making it well-suited for evaluating the model's performance in industrial settings. Figure~\ref{fig:sup_qual} shows an example from this dataset.

The \textit{unsupervised setting} is evaluated on MVTec AD~\citep{mvtec} and VisA~\citep{visa} datasets. 
\textbf{MVTec AD} comprises 15 categories, while \textbf{VisA} contains 12 different categories. Every category contains only normal images in the train set, intended to be used with unsupervised methods. The test set includes both normal and annotated anomalous images, covering a diverse range of anomaly types, scales, and complexities. Figure~\ref{fig:unsup_qual} shows some examples from both datasets.

\subsection{Evaluation metrics}

Dataset-specific evaluation metrics are adopted in line with original protocols and recent literature~\citep{fuvcka_transfusion,bozic2021end2end,racki_sensum}. Image-level performance for SensumSODF, MVTec AD, and VisA is measured using the Area Under the Receiver Operator Curve (AUROC). The Area Under the Per-Region Overlap (AUPRO) is used for pixel-level performance. 
For KSDD2, we follow the majority of related works by using Average Precision for both image-level detection and pixel-level localisation performance ($AP_{det}$ and $AP_{loc}$).

\subsection{Implementation details}
\label{exp:impl}

\textbf{Training} The model is trained for 300 epochs using the AdamW~\citep{loshchilov2018adamw} optimizer with a batch size of 32. Larger batch sizes help improve the results due to the greater variety of synthetic anomalies generated within the batch. The learning rate for the segmentation and classification head is set to $2\cdot10^{-4}$ with a weight decay of $10^{-5}$, while the learning rate of the adaptor module is set to $10^{-4}$.

A learning rate scheduler that reduces the learning rate by a factor of 0.4 after 240 and 270 epochs is used to stabilise the training. The gradient is clipped to norm 1 to further stabilise the training in the supervised setting. In contrast, in the unsupervised setting, we stop the gradient flow from the classification head to the segmentation head.

\textbf{Synthetic anomaly generation} Gaussian noise for synthetic anomalies is sampled from $\mathcal{N}(0, \sigma^2)$ with $\sigma=0.015$. The Perlin noise binarisation threshold varies according to the supervision regime. In the mixed and fully supervised setting, the pixel-level labels efficiently capture abnormality distribution, so synthetic anomalies primarily refine the normality boundary. To keep these anomalies small, a threshold of $0.6$ is used. Due to the lack of pixel-level labelled defects in the weakly supervised setting, synthetic anomalies are required to enhance the modelling of the global distribution. Larger synthetic anomalies are needed for this, achieved with a threshold of $0.2$. For unsupervised settings, thresholds used are: $0.6$ for VisA and $0.2$ for MVTec AD.

\textbf{Images}
All input images are normalised using ImageNet~\citep{deng_imagenet} normalisation. For MVTec AD and VisA, images are resized to $256 \times 256$, in line with recent literature~\citep{fuvcka_transfusion, zavrtanik_draem}. For SensumSODF and KSDD2, we follow the original protocols from~\cite{racki_sensum, KSDD2} and use $232 \times 640$ for KSDD2, $192 \times 320$ for SensumSODF capsule and $144 \times 144$ for SensumSODF softgel. All compared models use the same image sizes to ensure a fair comparison.

Flipping augmentations are applied in supervised settings to extend the set of anomalous images, as described in~\cite{racki_sensum}. To improve training stability, we follow~\cite{KSDD2} by balancing each epoch with an equal number of normal and anomalous samples.

\textbf{Evaluation}
To enable comparison with the base SimpleNet~\citep{liu_simplenet}, we modify its loss design to support fully supervised training by classifying defective samples in the ground truth mask ($\mathrm{M}_{gt}$) as anomalous. All other parameters stay consistent with those in the original work~\citep{liu_simplenet}.

Metrics are calculated using the model obtained from the final training epoch in all cases. For SimpleNet and SuperSimpleNet, we perform 5 training runs with different random seeds and report the mean and standard deviation of the results.
We use the predefined train-test splits for KSDD2, MVTec AD, and VisA, while we use 3-fold cross-validation for SensumSODF, as outlined in the original paper~\citep{racki_sensum}. The test sets in all cases contain both real normal and anomalous images. All of the compared models follow the same evaluation protocol. Above-described hyperparameters were initially adapted from SimpleNet and empirically evaluated for best performance. We include the hyperparameter sensitivity analysis performed by cross-validation on SensumSODF in the Supplementary material. We adhere to standard protocol~\citep{batzner_efficientad, roth_patchcore, KSDD2, racki_sensum} and fix the hyperparameters for all categories in datasets to enable fair comparison.

\subsection{Experimental results}
\label{sec:results}

\subsubsection{Results in the fully supervised setting.} 
SuperSimpleNet is compared with the current state-of-the-art methods for the fully supervised setting: SegDecNet~\citep{KSDD2}, TriNet~\citep{racki_sensum}, MaMiNet~\citep{luo2023maminet}, BGAD~\citep{yao_bgad}, PRN~\citep{zhang_prn}, and modified SimpleNet~\citep{liu_simplenet}. 

The results of anomaly detection and localisation on SensumSODF and KSDD2 datasets are presented in Table~\ref{tab:fully}.
SuperSimpleNet achieves the best results on SensumSODF, with a mean anomaly detection AUROC of $98.0~\%$, surpassing the previous state-of-the-art by $1.1$ percentage points (p.p.), reducing the error by 35.5~\%. 
SuperSimpleNet also achieves a state-of-the-art detection $\text{AP}_{\text{det}}$ of $97.8~\%$ on KSDD2, surpassing the previous best result by $1.6$ p.p., reducing the error by 42~\%. It demonstrates greater stability than SimpleNet on both datasets, as indicated by the lower reported standard deviations.

\begin{table}[t]
    \centering
    \setlength{\tabcolsep}{4pt}
    \begin{tabular}{lcccc}
    \toprule
         \multirow{2}{*}{~}&  \multicolumn{2}{c}{SensumSODF}&  \multicolumn{2}{c}{KSDD2}\\
         \cmidrule(lr){2-3}\cmidrule(lr){4-5}
         &  Det.&  Loc.&  Det.&Loc.\\ 
         \midrule
		SegDecNet~{\footnotesize \citep{KSDD2}}& 83.4& 75.2& 93.9& 75.0\\
		TriNet~{\footnotesize \citep{racki_sensum}}& 96.9 & - & - & -\\
		MaMiNet~{\footnotesize \citep{luo2023maminet}} & - & -& 96.2 & -\\
		DSR~{\footnotesize \citep{zavrtanik2022dsr}} & - & -& 95.2& 85.5\\
		PRN~{\footnotesize \citep{zhang_prn}}& 80.6& 66.0& 78.6& 48.5\\
		BGAD~{\footnotesize \citep{yao_bgad}}& 94.3& 97.0& 92.7& 76.5\\
		SimpleNet~{\footnotesize \citep{liu_simplenet}}& \meanwithstd{88.4}{1.84}& \meanwithstd{89.6}{1.14}& \meanwithstd{93.5}{1.05}& \meanwithstd{75.9}{2.40}\\
		\textbf{Ours}~& \meanwithstd{98.0}{0.19}& \meanwithstd{95.8}{0.28}& \meanwithstd{97.8}{0.18}& \meanwithstd{81.3}{0.64}\\

    \bottomrule
    \end{tabular}
    \caption{Results of fully supervised anomaly detection and localisation on the SensumSODF dataset (AUROC and AUPRO) and KSDD2 ($\text{AP}_{\text{det}}$ and $\text{AP}_{\text{loc}}$). }
    \label{tab:fully}
\end{table}

We hypothesise that the classification head's simple yet powerful design is the primary reason behind the high detection performance. It can efficiently learn to capture more global information in the presence of real and synthetic anomalies, enabling a more robust distinction between normal and anomalous samples.

\begin{figure*}[!h]
  \centering
   \includegraphics[width=1\linewidth]{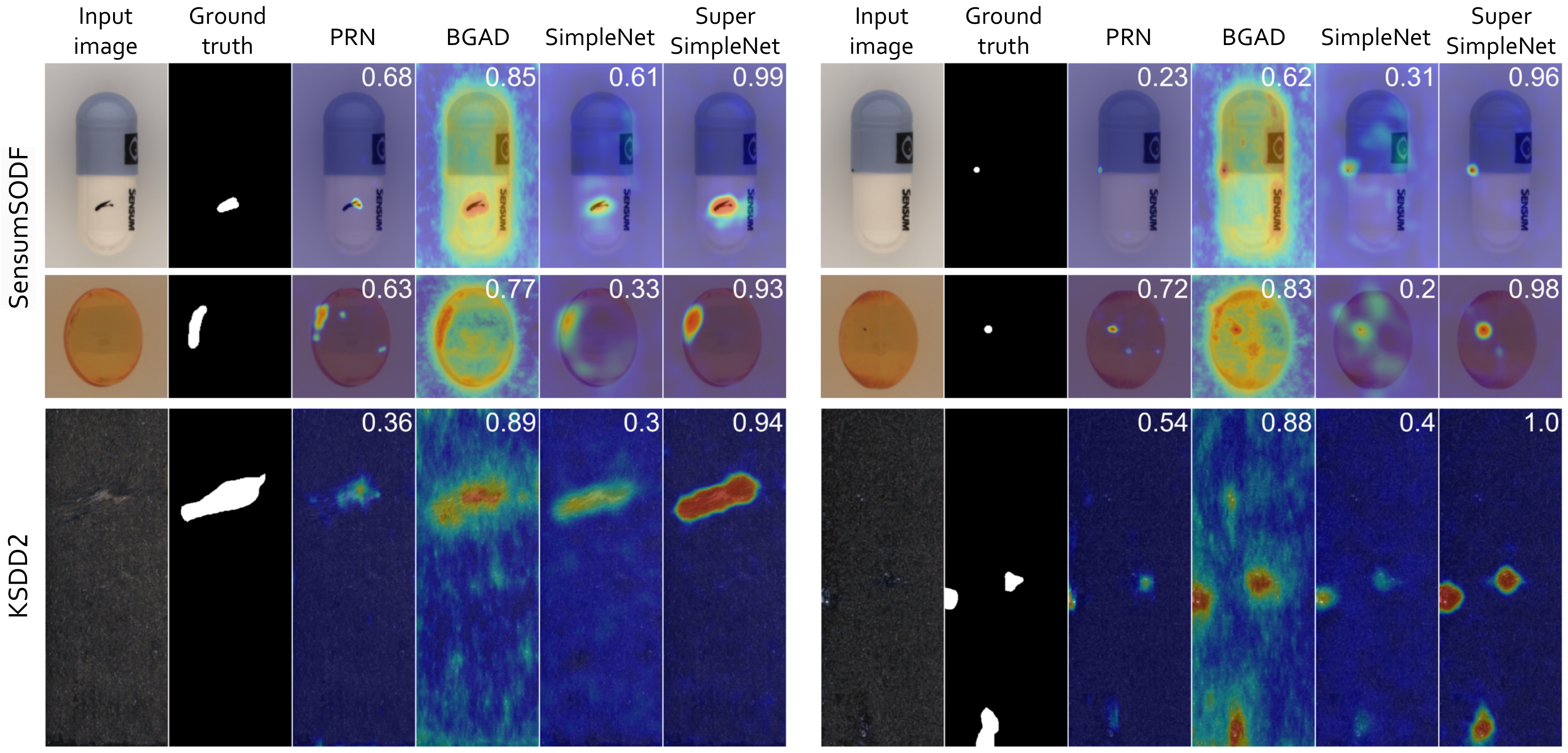}
   \caption{Qualitative comparison of anomaly maps produced in a fully supervised setting on SensumSODF and KSDD2. The first two rows display SensumSODF samples (capsule and softgel), while the last row shows KSDD2 examples. Each sample includes the input image, ground truth, and overlaid anomaly maps for each model. The anomaly scores are displayed in the top-right corner of each anomaly map. 
   }    
   \label{fig:sup_qual}
\end{figure*}

The qualitative examples for the fully supervised setting are shown in Figure~\ref{fig:sup_qual}. Compared to related methods, SuperSimpleNet predicts anomaly maps with less noise, as well as more accurate anomaly scores.

\subsubsection{Results in the weakly supervised setting.} 

Table~\ref{tab:weakly} displays results of weakly supervised setting for SensumSODF and KSDD2, compared against current state-of-the-art methods SegDecNet~\citep{KSDD2}, TriNet~\citep{racki_sensum}, MaMiNet~\citep{luo2023maminet}, and DRA~\citep{ding2022dra} that support weak supervision. 

\begin{table}[h]
    \centering
    \setlength{\tabcolsep}{4pt}
    \begin{tabular}{lccccc}
    \toprule
         \multirow{2}{*}{~}&  \multicolumn{2}{c}{SensumSODF}&  \multicolumn{2}{c}{KSDD2}\\
         \cmidrule(lr){2-3}\cmidrule(lr){4-5}
         &  Det.&  Loc.&  Det.&Loc.\\ 
         \midrule
		SegDecNet~{\footnotesize \citep{KSDD2}} & - & -& 73.3& 1.0\\
		TriNet~{\footnotesize \citep{racki_sensum}}& 91.5 & - & - & -\\
		MaMiNet~{\footnotesize \citep{luo2023maminet}} & - & -& 80.0 & -\\
		DRA~{\footnotesize \citep{ding2022dra}}& 90.1 & -& 89.3 & -\\
		\textbf{Ours}~& \meanwithstd{97.4}{0.11}& \meanwithstd{92.8}{2.12}& \meanwithstd{97.2}{0.48}& \meanwithstd{47.6}{2.46}\\

    \bottomrule
    \end{tabular}
    \caption{Results of weakly supervised anomaly detection and localisation on the SensumSODF dataset (AUROC and AUPRO), and KSDD2 (AP-det and AP-loc).}
    \label{tab:weakly}
\end{table}
Despite no pixel-level annotations, SuperSimpleNet achieves a high detection result of $97.4~\%$ AUROC on SensumSODF. This is only $0.6$ p.p lower than in a fully supervised scenario, while the performance of the best related method TriNet~\citep{racki_sensum} drops by $5.4$ p.p. Our method also obtains a localisation result of $92.8~\%$ AUPRO, while the other methods either don't produce localisation (DRA) or don't evaluate it (SegDecNet, TriNet and MaMiNet). 
On KSDD2, SuperSimpleNet achieves strong detection and localisation results with $97.2~\%~\text{AP}_{\text{det}}$ and $47.6~\%~\text{AP}_{\text{loc}}$. Our method again displays good performance in the absence of pixel-level labels, with only $0.6$ p.p. reduction compared to the fully supervised setting. In contrast, the performance of related methods like SegDecNet~\citep{KSDD2} and MaMiNet~\citep{luo2023maminet} drops by $22.1$ p.p and $16.2$ p.p., respectively. Compared to related methods, SuperSimpleNet also excels in localisation performance due to synthetic anomaly generation during training.

\subsubsection{Results using mixed supervision}

\begin{figure}[!h]
    \centering
    \includegraphics[width=1\linewidth]{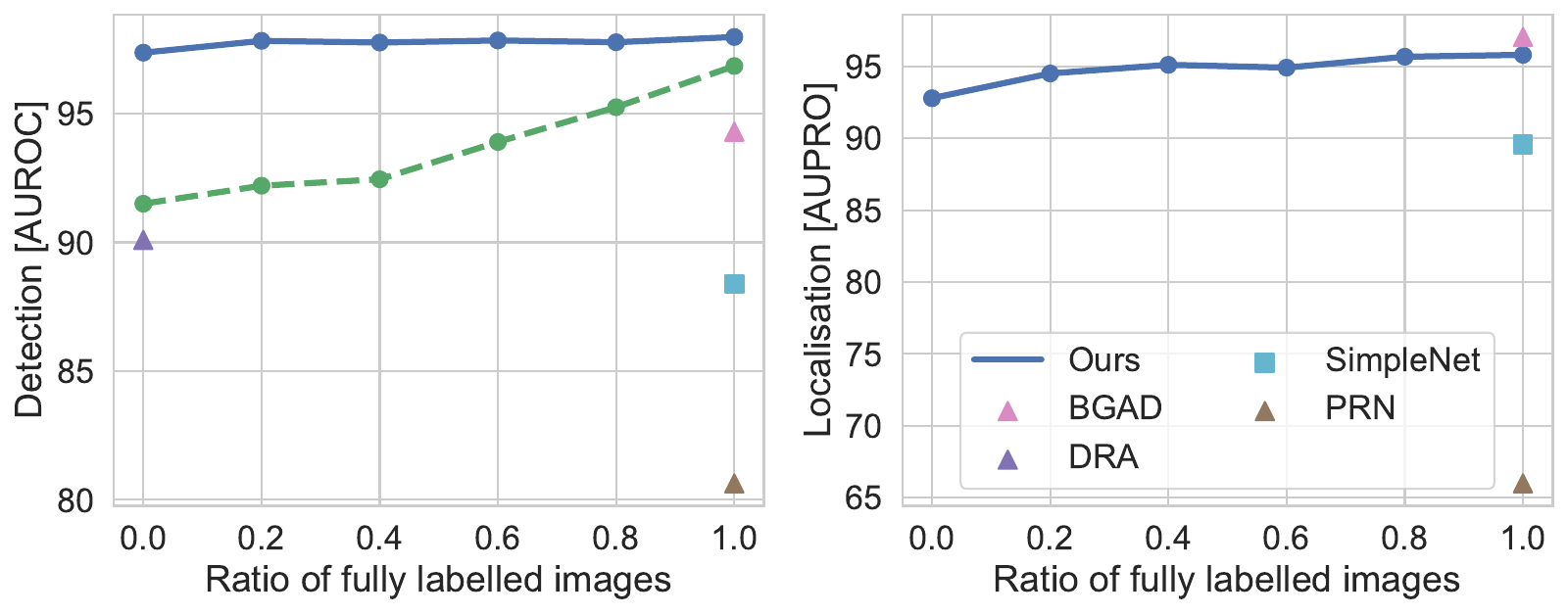}
    \caption{Results of anomaly detection (AUROC) and localisation (AUPRO) on the SensumSODF dataset using mixed supervision. The ratio of available pixel-level labels is displayed on the x-axis.}
    \label{fig:sensum_mixed}
\end{figure}

Figure~\ref{fig:sensum_mixed} showcases the results obtained using mixed supervision. Unlike previous methods, SuperSimpleNet relies less on pixel-level labels, as reflected by superior performance when fewer annotations are available. This allows SuperSimpleNet to outperform TriNet's~\citep{racki_sensum} fully supervised performance in weakly supervised mode.

\begin{figure}[t]
    \centering
    \includegraphics[width=1\linewidth]{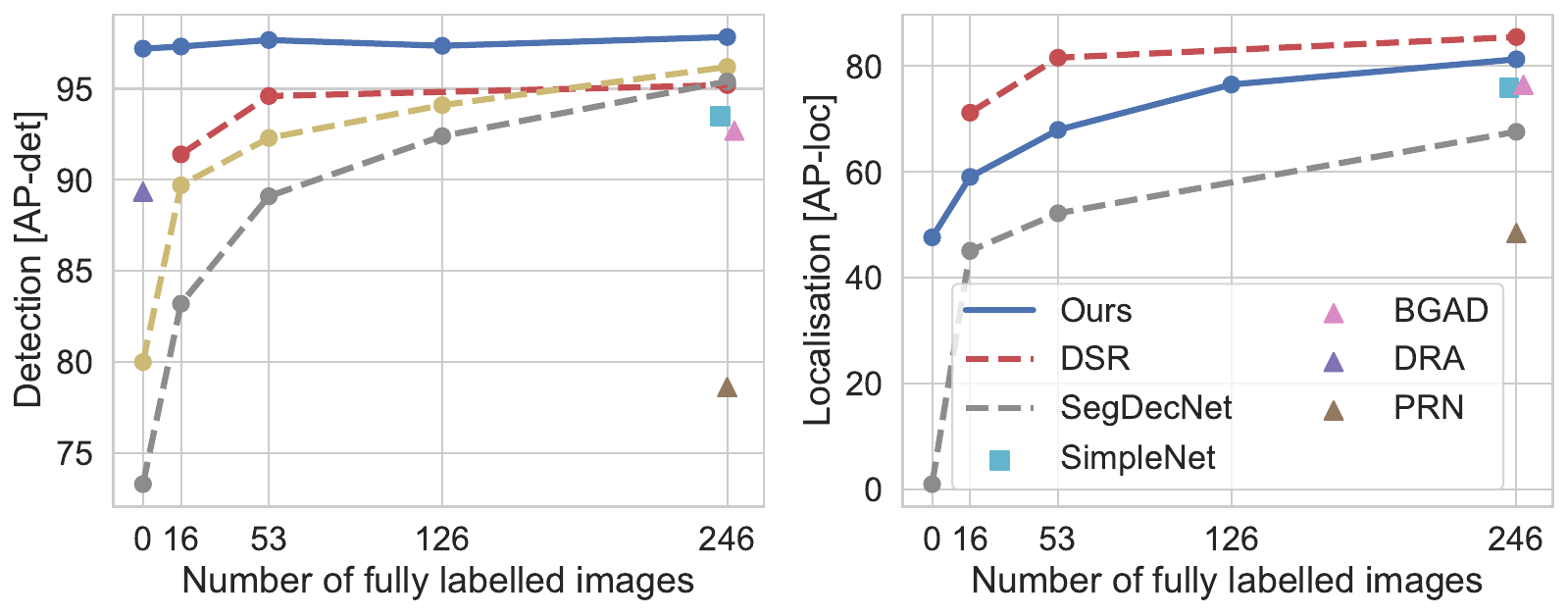}
    \caption{Results for anomaly detection ($\text{AP}_{\text{det}}$) and localisation ($\text{AP}_{\text{loc}}$) on the KSDD2 dataset using mixed supervision. The number of pixel-level labels is displayed on the x-axis.}
    \label{fig:ksdd_mixed}
\end{figure}

Performance using mixed supervision on KSDD2 is displayed in Figure~\ref{fig:ksdd_mixed}. In this case, results of DSR~\citep{zavrtanik2022dsr} are also reported, where the number of labelled samples indicates the number of fully-labelled anomalous samples used, while the method is unable to utilize the remaining weakly labelled samples.

SuperSimpleNet outperforms related methods in detection, even with all samples only weakly labelled. Other methods require a significant number of fully labelled samples to reach performance comparable to that of a fully supervised mode. This, in practice, translates into a substantial reduction in the labelling efforts, making SuperSimpleNet highly suitable for real-world industrial environments. These results also prompt future research to focus on methods capable of mixed supervision.

While detection is the most important task for industrial scenarios, SuperSimpleNet also outperforms SegDecNet~\citep{KSDD2} in localisation, thanks to the generated synthetic anomalies. However, DSR does outperform SuperSimpleNet in the localisation task.

\begin{figure}[!h]
    \centering
    \includegraphics[width=1\linewidth]{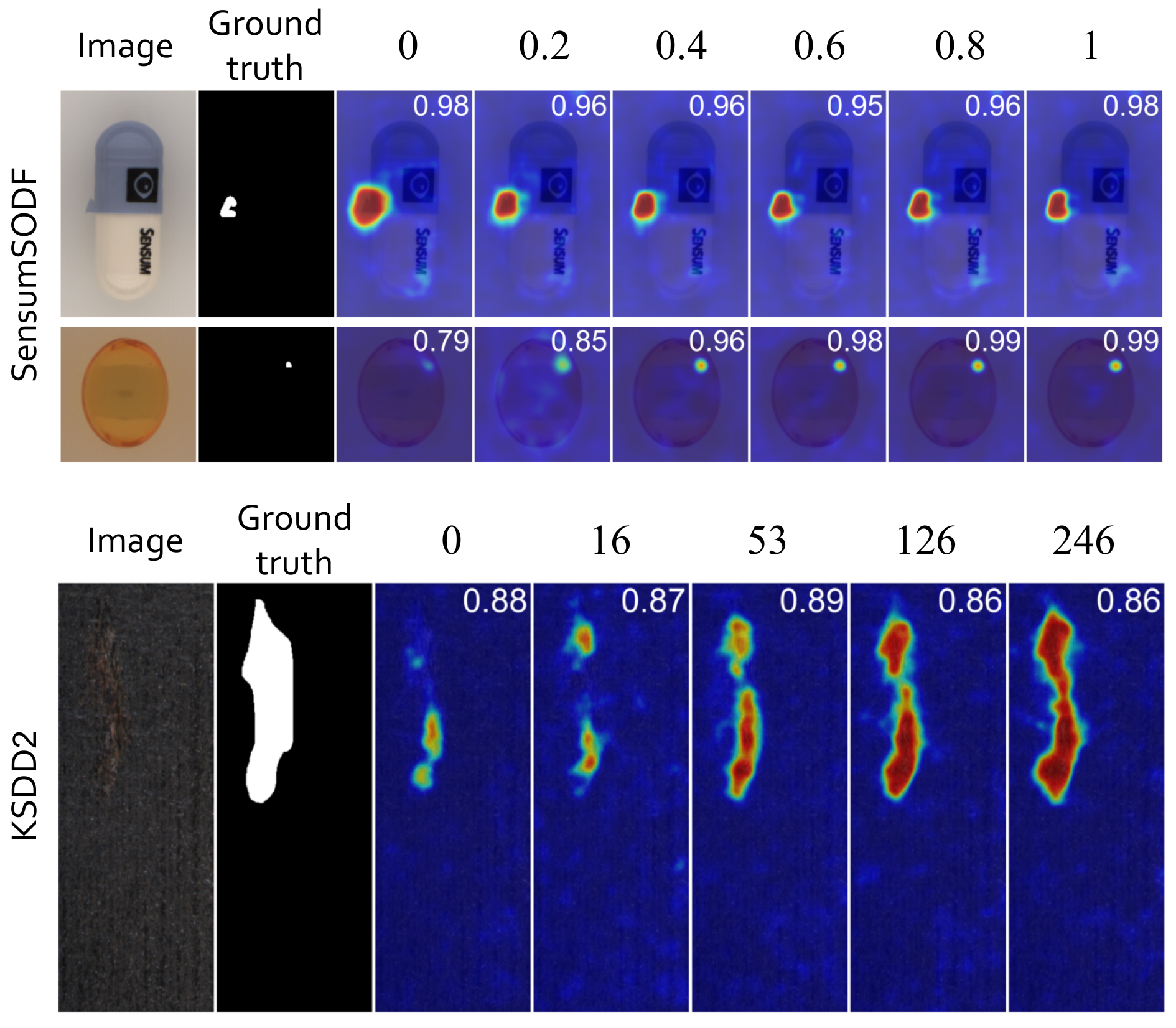}
    \caption{Qualitative comparison of anomaly maps produced in the mixed supervision setting on SensumSODF and KSDD2. The input image, the ground truth, and the overlaid anomaly maps for different ratio (or number) of fully labelled images are shown. The anomaly score is shown in the top right corner of each map.}
    \label{fig:qual_mixed}
\end{figure}

Figure~\ref{fig:qual_mixed} presents qualitative results on SensumSODF and KSDD2. The localisation performance noticeably improves with the availability of more labels, while anomaly scores are already well predicted without any pixel-level labels. This observation aligns with the previously discussed quantitative results.

We also include results in a numeric tabular format in the Supplementary material. 

\subsubsection{Results in the unsupervised setting.} 
SuperSimpleNet is compared with the current state-of-the-art methods for the unsupervised setting: AST~\citep{rudolph_ast}, DSR~\citep{zavrtanik2022dsr}, EfficientAD~\citep{batzner_efficientad}, FastFlow~\citep{yu_fastflow}, Patchcore~\citep{roth_patchcore}, DR{\AE}M~\citep{zavrtanik_draem} and SimpleNet~\citep{liu_simplenet}. 

Table~\ref{tab:unsup} presents the results on the MVTec AD and VisA datasets.
SuperSimpleNet achieves state-of-the-art performance with mean anomaly detection of $98.3\%$ on MVTec AD and anomaly detection AUROC of $93.6\%$ on VisA. 
While SuperSimpleNet in this setting doesn't perform as efficiently as in supervised settings, it still achieves great results, outperforming the original SimpleNet in anomaly detection and demonstrating more stable results overall.

\begin{table}[h]
    \centering
    \setlength{\tabcolsep}{4pt}
    \begin{tabular}{lcccc}
    \toprule
         \multirow{2}{*}{~}&  \multicolumn{2}{c}{MVTec AD}&  \multicolumn{2}{c}{VisA}\\
         \cmidrule(lr){2-3}\cmidrule(lr){4-5}
         &  Det.&  Loc.&  Det.&Loc.\\ 
         \midrule
		AST~{\footnotesize \citep{rudolph_ast}}& 98.9& 81.2& 94.9& 81.5\\
		DSR~{\footnotesize \citep{zavrtanik2022dsr}}& 98.1& 90.8& 91.8& 68.1\\
		EfficientAD~{\footnotesize \citep{batzner_efficientad}}& 99.1& 93.5& 98.1& 94.0\\
		FastFlow~{\footnotesize \citep{yu_fastflow}}& 96.9& 92.5& 93.9& 86.8\\
		PatchCore~{\footnotesize \citep{roth_patchcore}}& 98.7& 92.7& 94.3& 79.7\\
		DRÆM~{\footnotesize \citep{zavrtanik_draem}}& 98.0& 92.8& 91.5& 78.0\\
		SimpleNet~{\footnotesize \citep{liu_simplenet}}& \meanwithstd{97.6}{0.40}& \meanwithstd{90.5}{0.75}& \meanwithstd{91.2}{1.08}& \meanwithstd{88.0}{0.87}\\
		\textbf{Ours}~& \meanwithstd{98.3}{0.14}& \meanwithstd{91.2}{0.14}& \meanwithstd{93.6}{0.77}& \meanwithstd{87.4}{0.98}\\

    \bottomrule
    \end{tabular}
    \caption{Anomaly detection and localisation (AUROC and AUPRO) on MVTec AD and VisA datasets.}
    \label{tab:unsup}
\end{table}

\begin{figure*}[!h]
  \centering
   \includegraphics[width=1\linewidth]{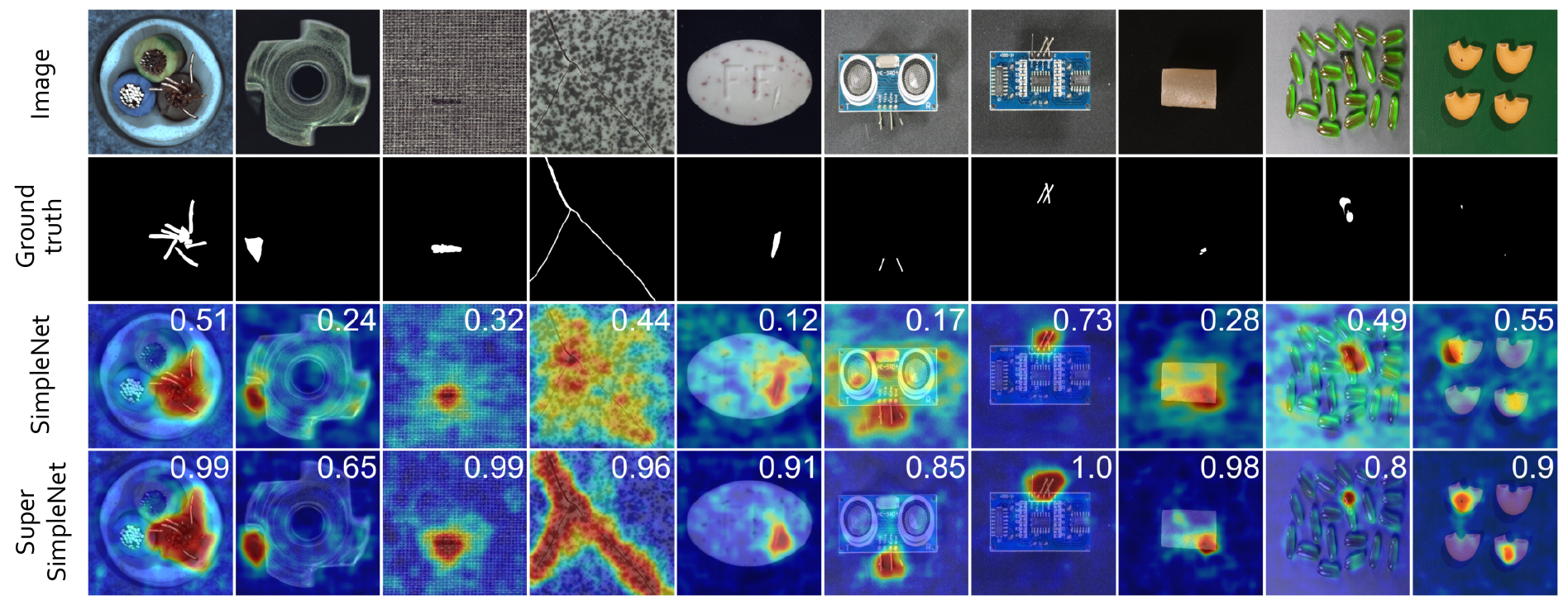}
   \caption{Qualitative comparison of anomaly maps produced by unsupervised SuperSimpleNet and SimpleNet. The top row presents the input image, followed by the ground truth anomaly mask in the second row. The third and fourth rows display the anomaly maps produced by SimpleNet and SuperSimpleNet, respectively. The anomaly score is indicated in the top right corner of each anomaly map. 
   }    
   \label{fig:unsup_qual}
\end{figure*}

\noindent Qualitative results of the unsupervised model on MVTec AD~\citep{mvtec} and VisA~\citep{visa} datasets are presented in Figure~\ref{fig:unsup_qual}. 
SuperSimpleNet generates more accurate anomaly scores than SimpleNet. The anomaly maps are also easier to interpret due to the better separation of high-certainty defective areas from noise.

\subsubsection{Computational efficiency.} 

\begin{figure}[!h]
    \centering
    \includegraphics[width=1\linewidth]{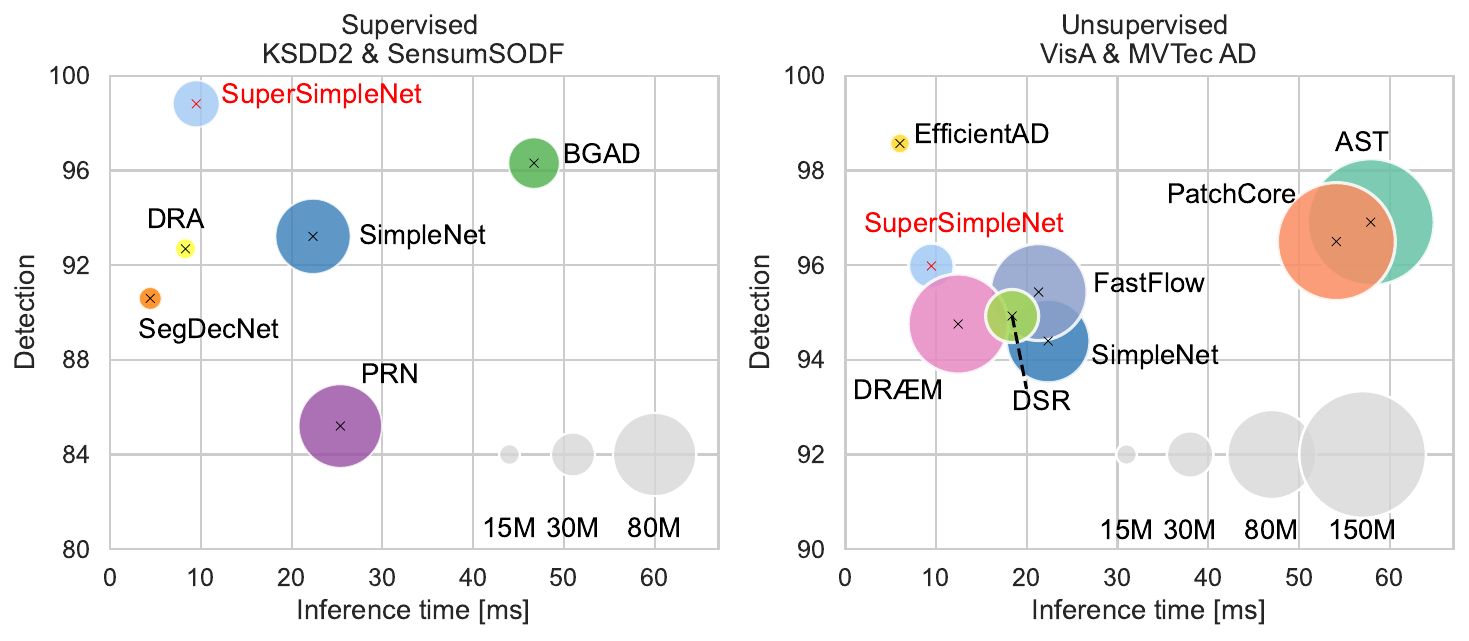}
    \caption{Inference time (\textbf{milliseconds} - lower is better) and anomaly detection performance (\textbf{AUROC} - higher is better) for different models, measured on an NVIDIA Tesla V100S. The size of the circles represents the model's parameter size.}
    \label{fig:speed}
\end{figure}

Figure~\ref{fig:speed} showcases computational efficiency results, measured using an NVIDIA Tesla V100S following the benchmark protocol introduced in EfficientAD~\citep{batzner_efficientad}. SuperSimpleNet achieves an inference time of 9.5~ms and a throughput of 262 images per second, making it the fastest method that works in both unsupervised and supervised settings.

\subsection{Ablation Study}
\label{sec:ablation}

In this section, we determine the contribution of each SuperSimpleNet module, investigate its performance in the medical domain, analyse failure cases, and validate performance at higher resolutions. First, we evaluate our contributed modules by systematically removing each component from the architecture and analysing the impact on performance. The results are shown in Table~\ref{tab:components}. For comparison, we also include the results of the original SimpleNet~\citep{liu_simplenet} and a version of it with the new synthetic generation strategy.

\begin{table*}[!h]
    \centering
    \setlength{\tabcolsep}{3pt}
    \resizebox{\linewidth}{!}{
    \begin{tabular}{lcccccccccccc}
            \toprule
        \multirow{2}{*}{\textbf{Method}}~ &
        \multicolumn{2}{c}{Synth. anom.} &
        \multicolumn{4}{c}{Architecture} &
        \multicolumn{2}{c}{\textit{Fully Sup.}} &
        \multicolumn{2}{c}{\textit{Unsup.}} & 
        \multicolumn{2}{c}{\textit{Avg.}}\\
                \cmidrule(lr){2-3}\cmidrule(lr){4-7}\cmidrule(lr){8-9}\cmidrule(lr){10-11}\cmidrule(lr){12-13}
         ~ & SSN & SN & Upscale & \makecell{Cls. \\ head} & \makecell{Train \\ opti.} & \makecell{Loss \\ Weight} & Det.  & Loc.  & Det.  & Loc. & Det.  & Loc.\\ \midrule
    \rowcolor{gray!20} \textbf{Ours} &\checkmark &~ &\checkmark &Simple &\checkmark &\checkmark &98.8& 97.3& 96.0& 89.3& 97.4& 93.3\\ 
$SSN_{no\_upscale}$ &\checkmark &~ &~ &Simple &\checkmark &\checkmark &\textcolor{red}{-0.7}& \textcolor{red}{-1.5}& \textcolor{red}{-0.7}& \textcolor{red}{-1.3}& \textcolor{red}{-0.7}& \textcolor{red}{-1.4}\\ 
$SSN_{no\_cls}$ &\checkmark &~ &\checkmark &~ &\checkmark &\checkmark &\textcolor{red}{-1.0}& \textcolor{blue}{+0.2}& \textcolor{blue}{+0.1}& \textcolor{blue}{+0.1}& \textcolor{red}{-0.5}& \textcolor{blue}{+0.1}\\ 
$SSN_{complex\_cls}$ &\checkmark &~ &\checkmark &Complex &\checkmark &\checkmark &\textcolor{gray}{0.0}& \textcolor{blue}{+0.2}& \textcolor{red}{-2.6}& \textcolor{red}{-0.7}& \textcolor{red}{-1.3}& \textcolor{red}{-0.2}\\ 
$SSN_{cls\_no\_Mo}$ &\checkmark &~ &\checkmark &No $\text{M}_o$ &\checkmark &\checkmark &\textcolor{red}{-0.4}& \textcolor{blue}{+0.2}& \textcolor{red}{-46.0}& \textcolor{red}{-1.8}& \textcolor{red}{-23.2}& \textcolor{red}{-0.8}\\ 
$SSN_{cls\_w\_adapt}$ &\checkmark &~ &\checkmark &$\mathcal{A}$ in cls &\checkmark &\checkmark &\textcolor{red}{-0.2}& \textcolor{blue}{+0.1}& \textcolor{red}{-0.3}& \textcolor{gray}{0.0}& \textcolor{red}{-0.3}& \textcolor{gray}{0.0}\\ 
$SSN_{old\_train}$ &\checkmark &~ &\checkmark &Simple &~ &\checkmark &\textcolor{red}{-1.4}& \textcolor{red}{-1.8}& \textcolor{red}{-1.8}& \textcolor{red}{-5.4}& \textcolor{red}{-1.6}& \textcolor{red}{-3.6}\\ 
$SSN_{overlap}$ &Overlap &~ &\checkmark &Simple &\checkmark &\checkmark &\textcolor{red}{-0.2}& \textcolor{gray}{0.0}& \textcolor{gray}{0.0}& \textcolor{gray}{0.0}& \textcolor{red}{-0.1}& \textcolor{gray}{0.0}\\ 
$SSN_{no\_loss\_weight}$ &\checkmark &~ &\checkmark &Simple &\checkmark &~ &\textcolor{red}{-0.4}& \textcolor{red}{-1.9}& \textcolor{gray}{0.0}& \textcolor{gray}{0.0}& \textcolor{red}{-0.2}& \textcolor{red}{-0.9}\\ 
$SSN_{SN\_anom}$ &~ &\checkmark &\checkmark &Simple &\checkmark &\checkmark &\textcolor{red}{-0.4}& \textcolor{blue}{+0.7}& \textcolor{red}{-4.4}& \textcolor{red}{-1.0}& \textcolor{red}{-2.4}& \textcolor{red}{-0.1}\\ 
$SSN_{no\_cls\&SN\_anom}$ &~ &\checkmark &\checkmark &~ &\checkmark &\checkmark &\textcolor{red}{-1.3}& \textcolor{blue}{+0.9}& \textcolor{gray}{0.0}& \textcolor{red}{-0.6}& \textcolor{red}{-0.7}& \textcolor{blue}{+0.2}\\ 
$SSN_{complex\_cls\&SN\_anom}$ &~ &\checkmark &\checkmark &Complex &\checkmark &\checkmark &\textcolor{red}{-0.5}& \textcolor{blue}{+0.7}& \textcolor{red}{-29.6}& \textcolor{red}{-2.9}& \textcolor{red}{-15.0}& \textcolor{red}{-1.1}\\ 
$SSN_{no\_anom}$ &~ &~ &\checkmark &Simple &\checkmark &\checkmark &\textcolor{red}{-0.3}& \textcolor{red}{-2.5}& -& -& - & -\\ 

\hline\hline
$SN$ & ~ & \checkmark & ~ & ~& ~ & ~ & \textcolor{red}{-5.6}& \textcolor{red}{-4.1}& \textcolor{red}{-1.6}& \textcolor{gray}{0.0}& \textcolor{red}{-3.6}& \textcolor{red}{-2.1}\\ 
$SN_{SSN\_anom}$ & \checkmark &  ~& ~ & ~ & ~ & ~ &\textcolor{red}{-7.5}& \textcolor{red}{-6.3}& \textcolor{red}{-1.4}& \textcolor{red}{-3.1}& \textcolor{red}{-4.4}& \textcolor{red}{-4.7}\\ 

    \end{tabular}
    }
    \caption{Ablation study results on anomaly detection and localisation (AUROC / AUPRO) in the fully supervised setting (average of results on SensumSODF and KSDD2) and the unsupervised setting (average of results on MVTec AD and VisA), as well as the average of both settings. $SSN$ stands for SuperSimpleNet, while $SN$ stands for SimpleNet. Every change in performance is also reflected in colour: \textcolor{blue}{blue} indicates better result, \textcolor{red}{red} indicates worse result, and \textcolor{gray}{gray} indicates no change in result.}
    \label{tab:components}
\end{table*}

\noindent\textbf{Upscaling module.} 

Removing feature upscaling from the architecture ($SSN_{no\_upscale}$) leads to a reduction in overall detection performance by 0.7 p.p. The effect on localisation is even more pronounced, with a 1.4 p.p. reduction. These findings indicate the importance of feature scaling for achieving accurate final predictions. 

\noindent\textbf{Classification head.} 

The importance of the classification head is evident from the $SSN_{no\_cls}$ experiment, where the anomaly score $s$ is derived as the maximum value of the predicted anomaly map. This modification results in a decrease in supervised detection results by 1.0 p.p., as the removal of the classification head diminishes the network's discriminative power. On the other hand, this omission yields a slight improvement in the unsupervised setting. We hypothesise that reducing discriminative power helps mitigate overfitting to synthetic anomalies in this case.

\noindent\textbf{Simplicity of classification head.} 

An important part of SuperSimpleNet is a powerful but simple classification head. When we replace it with a more complex architecture consisting of three convolutional blocks, we obtain the results in $SSN_{complex\_cls}$ line. Results in the supervised detection remain unchanged, but unsupervised detection performance reduces by 2.6 p.p. These observations highlight the important role that the simplicity of the architecture plays in SuperSimpleNet, which still offers strong discriminative powers but simultaneously prevents overfitting.

\noindent\textbf{Classification head inputs.}

We remove the anomaly map ($\mathrm{M}_o$ predicted by the segmentation head) as a classification head input in the $SSN_{cls\_no\_Mo}$ experiment, forcing each head to learn independently. In the supervised setting, the performance slightly decreases (0.4 p.p.), suggesting that SuperSimpleNet does not heavily rely on pixel-level labels if real anomalies are available. However, the results in unsupervised settings significantly degrade, indicating that the anomaly map is crucial for preventing the classification head from overfitting to synthetic anomalies.

We also evaluate the impact of using adapted features $\mathcal{A}$ instead of regular features $\mathcal{F}$ in the classification head, as seen in the $SSN_{cls\_w\_adapt}$ experiment. This substitution results in a slight but non-negligible drop in detection performance across both settings. 

\noindent\textbf{Improved training.} 

The $SSN_{old\_train}$ experiment showcases the impact of upgrading the loss function, adding a learning rate scheduler, and adjusting gradient flow. These changes have a noticeable impact on results, as their absence leads to an overall 1.6 p.p. reduction in detection performance and a 3.6 p.p. decrease in localisation performance. We attribute this to more stable training in the final stages.

\noindent\textbf{Addition of synthetic anomalies to anomalous regions.}

We generate synthetic anomalies exclusively on non-anomalous regions when the image is fully labelled. The $SSN_{overlap}$ experiment shows the results where we remove this constraint. This leads to a slight reduction of 0.2 p.p. in detection performance, indicating that retaining as much genuine defect information as possible yields better performance. Unsupervised results remain unchanged, as this modification doesn't change anything in that case.

\noindent\textbf{Segmentation loss weighting.}
As described in Section~\ref{sec:loss}, we use segmentation loss weighting, proposed in~\cite{KSDD2}. As evident from line $SSN_{no\_loss\_weight}$, this approach contributes to detection performance (0.4 p.p.) and to localisation performance (1.9 p.p.). Given its simplicity, this technique could be widely adopted to enhance defect detection performance across supervised methods.

\noindent\textbf{Anomaly mask generation.} 

To verify the importance of the new anomaly mask generation strategy, we replace it with the original SimpleNet method, which involves perturbing the entire feature map with Gaussian noise.
The results from the $SSN_{SN\_anom}$ experiment reveal a reduction of 0.4 p.p. in the supervised setting and a larger decline of 4.4 p.p. in the unsupervised setting. We hypothesise that this drop in performance stems from the incompatibility of this strategy with SuperSimpleNet's classification head.

We further confirm this with $SSN_{no\_cls\&SN\_anom}$ experiment, where we apply SimpleNet's strategy without the classification head. Here, we do not observe this decline in the unsupervised setting. However, the performance in the supervised setting decreases by 1.3 p.p. 

Next, we combine SimpleNet's strategy with a complex classification head in the $SSN_{complex\_cls\&SN\_anom}$ experiment. The performance drop is larger than with a simple classification head (29.6 p.p.). This indicates that the complex classification head tends to overfit in the case of SimpleNet's strategy, further reinforcing the efficiency of SuperSimpleNet's architectural simplicity and anomaly mask generation strategy.

\begin{figure*}
    \centering
    \includegraphics[width=1\linewidth]{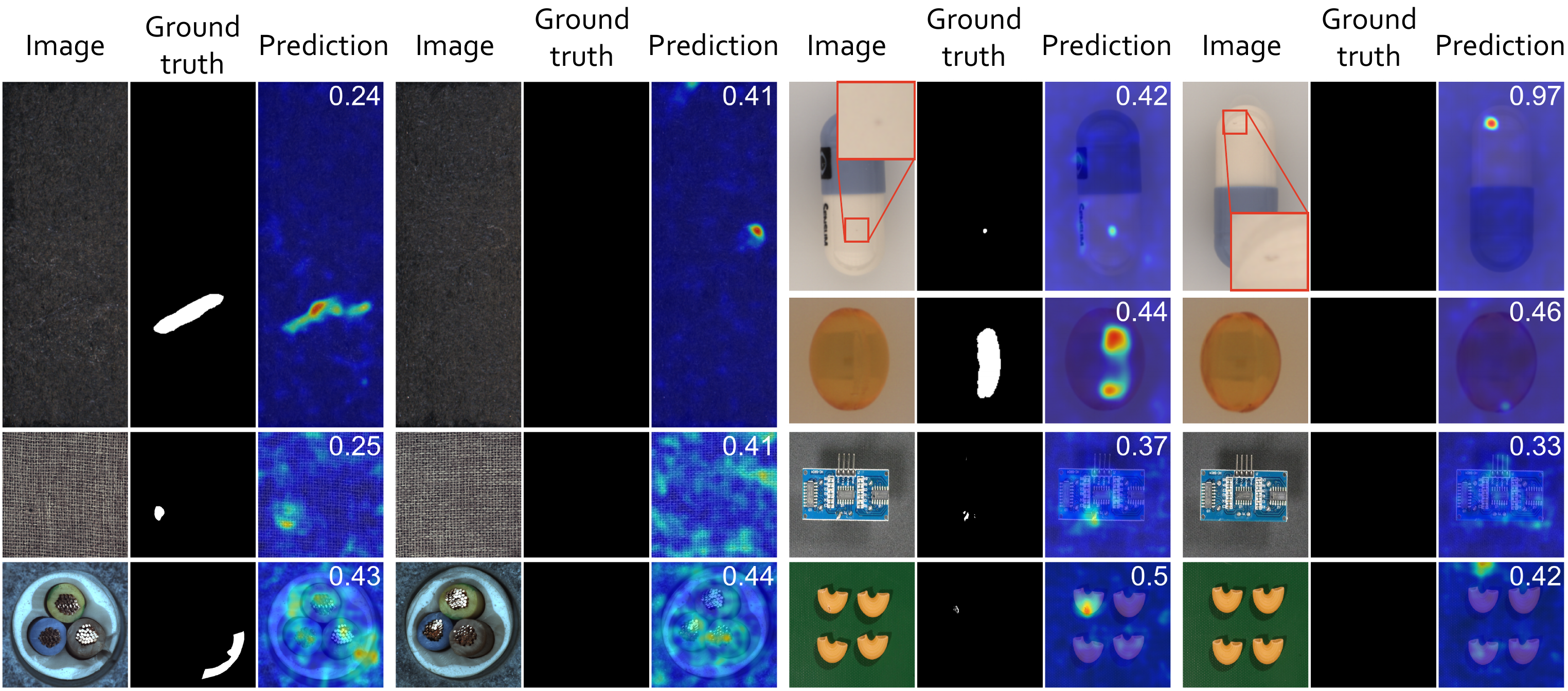}
    \caption{Failure cases with misclassified or low certainty predictions for KSDD2 (top left), SensumSODF (top two rows right), MVTec AD (bottom two rows left), and VisA (bottom two rows right). Each sample includes the input image, ground truth, and overlaid anomaly map with the anomaly scores displayed in the top-right corner.}
    \label{fig:fail_qual}
\end{figure*}

\noindent\textbf{Synthetic anomaly generation strategy.} 

The significance of synthetic anomalies in supervised training is highlighted in the $SSN_{no\_anom}$ experiment, where only real anomalies were used. The results demonstrate that including synthetic anomalies, even in the presence of real defects, improves detection performance by 0.3 p.p. and localisation performance by 2.5 p.p. It is also important to recognise that synthetic anomalies are the primary mechanism for unsupervised learning and segmentation training in weakly supervised settings. We do not report unsupervised results in this case, as the network fails to learn without synthetic anomalies. 

\noindent\textbf{Failure Cases.}

Figure~\ref{fig:fail_qual} depicts failure cases of SuperSimpleNet. The Figure illustrates that many of the undetected defects are challenging to spot because they closely resemble normal appearances. Additionally, we notice that some false positive predictions (VisA in the bottom right corner) result from debris in the background, which could be seen as an anomaly in some scenarios.

\noindent\textbf{Label ablation.}

\begin{figure}[h]
    \centering
    \includegraphics[width=1\linewidth]{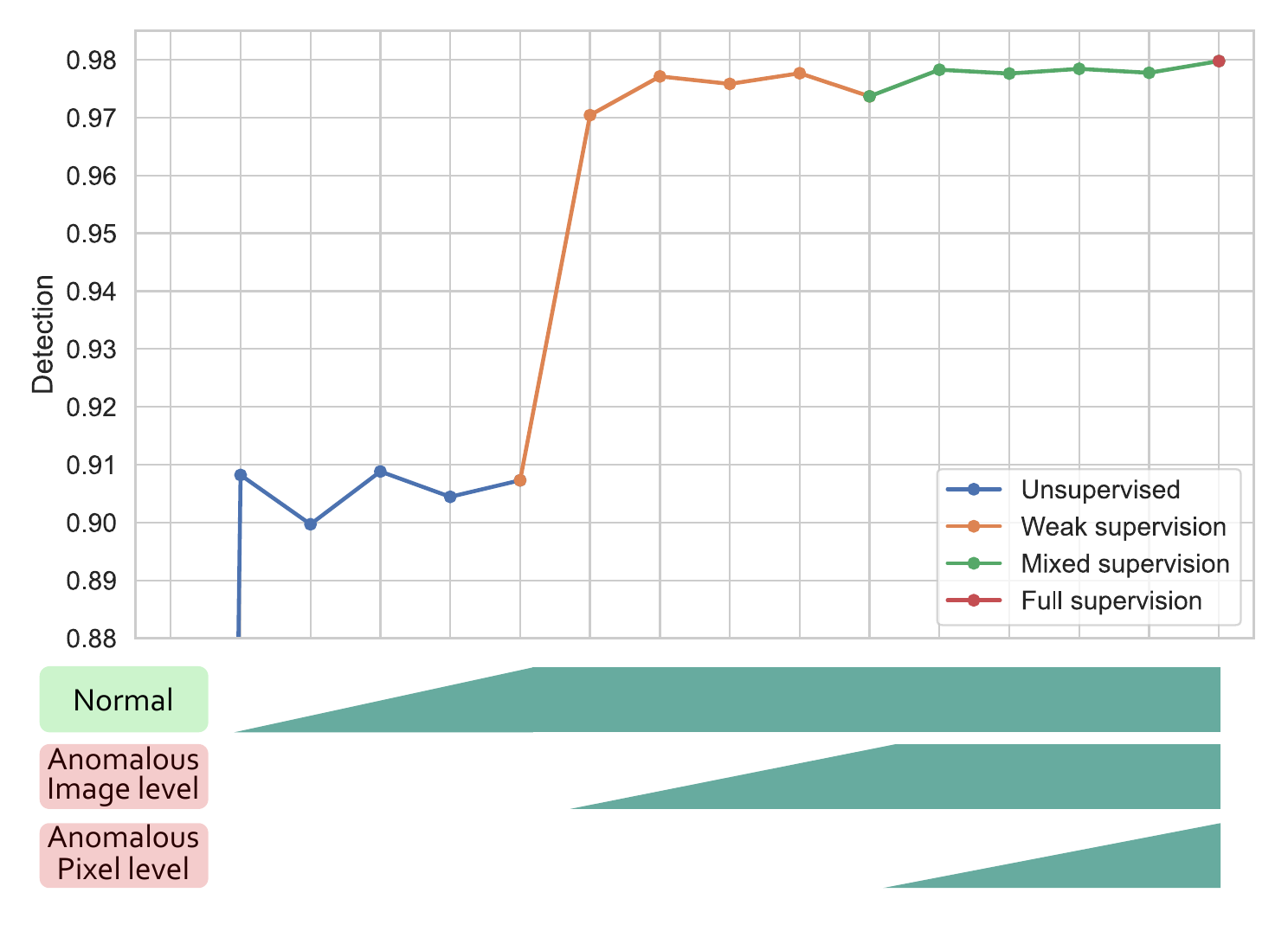}
    \caption{Results of label ablation study. The anomaly detection performance (AUROC) is shown with respect to the amount of labelled data. Each data regime corresponds to a specific supervision paradigm. Results indicate that our method is effective in resource-limited scenarios but enables improvement with additional annotated samples.}
    \label{fig:label_ablation}
\end{figure}

\begin{table*}[!h]
    \centering
    \small
    \setlength{\tabcolsep}{4pt}
    \begin{tabular}{lcccccccccc}
            \toprule
        \multirow{2}{*}{\textbf{Backbone}} &
        Inference Time & Memory & Param. &
        \multicolumn{2}{c}{\textit{Sup.}} &
        \multicolumn{2}{c}{\textit{Unsup.}} & 
        \multicolumn{2}{c}{\textit{Avg.}}\\
        
         ~ & [ms] & [MB] & [M] & Det.  & Loc.  & Det.  & Loc. & Det.  & Loc. &\\ \midrule
    \rowcolor{gray!20} WideResNet50 \textbf{Ours} &9.5 $\pm$ 0.1 & 413.1 $\pm$ 48.5 & 33.7 &98.8& 97.3& 96.0& 89.3& 97.4& 93.3\\ 
WideResNet100 &18.5 $\pm$ 0.4 & 565.2 $\pm$ 71.3 & 91.7 &\textcolor{gray}{0.0}& \textcolor{blue}{+0.3}& \textcolor{red}{-3.5}& \textcolor{red}{-3.7}& \textcolor{red}{-1.8}& \textcolor{red}{-1.7}\\ 
ResNet50 &10.4 $\pm$ 0.1 & 332.0 $\pm$ 8.0 & 17.4 &\textcolor{red}{-0.3}& \textcolor{gray}{0.0}& \textcolor{red}{-0.5}& \textcolor{red}{-0.3}& \textcolor{red}{-0.4}& \textcolor{red}{-0.1}\\ 
ResNet101 &18.3 $\pm$ 0.1 & 413.1 $\pm$ 0.0 & 36.4 &\textcolor{red}{-0.7}& \textcolor{blue}{+0.1}& \textcolor{red}{-1.5}& \textcolor{red}{-2.2}& \textcolor{red}{-1.1}& \textcolor{red}{-1.1}\\ 

\bottomrule
    \end{tabular}
    \caption{Average Anomaly Detection and Localisation results across the supervised and unsupervised training regimes for different feature extractors. In each row, the difference from the baseline model is shown. Performance deterioration is colored in \textcolor{red}{red}, performance boost is colored in \textcolor{blue}{blue}, and no changes are colored in \textcolor{gray}{gray}.}
    \label{tab:backbone}
\end{table*}

To assess the effect of the quantity of available data on the performance, an experiment is conducted on the SensumSODF Softgel category. First, the number of normal samples is increased, representing unsupervised learning with varying amounts of data. Next, keeping all the normal samples fixed, anomalous samples with image-level labels are progressively added, simulating weakly supervised learning with different amounts of anomalous data. Once all normal and anomalous images are included in the training set, pixel-level labels are added for the anomalous samples. This corresponds to learning with mixed supervision. Finally, the setting becomes fully supervised when all samples have pixel-level annotations.

The results depicted in Figure~\ref{fig:label_ablation} show that SuperSimpleNet already performs well with a limited amount of normal samples in the unsupervised setting. Adding just a few anomalous samples with image-level labels significantly boosts performance. However, the best results are only achieved when all the anomalous data has accompanying pixel-level annotations. This clearly demonstrates that SuperSimpleNet consistently improves with every additional available label (either image-level or pixel-level).

\noindent\textbf{Backbone ablation.}

To asses the importance of the chosen feature extractor, we have evaluated our model with several different and standard backbones. The results are presented in Table~\ref{tab:backbone} and show that WideResNet50 offers the best performance and efficiency.

\noindent\textbf{Generalisation to medical domain.}

We evaluate the generalisation capabilities of our method to the medical domain. The Table~\ref{tab:medical} contains results on medical unsupervised anomaly detection, including histopathology slides and brain MRI images, using the BMAD benchmark~\citep{bao2024bmad}. This showcases that SuperSimpleNet also works outside the scope of its development, that is, industrial inspection, and even further strengthens its contribution.

\begin{table}[!h]
    \centering
    \begin{tabular}{lcccc}
    \toprule
         &Brain MRI & Histopathology\\
         \midrule
		Padim~{\footnotesize \citep{defard_padim}}& \meanwithstd{79.0}{0.38}& \meanwithstd{67.2}{0.32}\\
		CFlow~{\footnotesize \citep{gudovskiy_cflow}}& \meanwithstd{74.8}{5.32}& \meanwithstd{55.7}{1.90}\\
		PatchCore~{\footnotesize \citep{roth_patchcore}}& \meanwithstd{91.6}{0.36}& \meanwithstd{69.3}{0.21}\\
		DRÆM~{\footnotesize \citep{zavrtanik_draem}}& \meanwithstd{62.4}{9.03}& \meanwithstd{52.3}{0.70}\\
		SimpleNet~{\footnotesize \citep{liu_simplenet}}& \meanwithstd{82.5}{3.34}& \meanwithstd{62.4}{3.71}\\
		\textbf{Ours}~& \meanwithstd{83.0}{2.86}& \meanwithstd{68.7}{4.56}\\

    \bottomrule
    \end{tabular}
    \caption{Results of unsupervised anomaly detection (AUROC) on the Brain MRI, Liver CT and Histopathology datasets.}
    \label{tab:medical}
\end{table}

\noindent\textbf{Image resolution ablation.}

In some industrial scenarios, it is required that the model operates with images of a higher resolution. This, however, decreases the computational efficiency of the model. While we have implicitly shown that SuperSimpleNet is capable of handling high-resolution images via KSDD2 and SensumSODF, where full-image resolution is used, we have decided to evaluate this tradeoff on VisA, where images of size $256 \times 256$ are used by default~\citep{fuvcka_transfusion,batzner_efficientad} to enable fair comparison. To assess this tradeoff, the resolution has been increased to $512 \times 512$. The results in Table~\ref{tab:det_res} show that higher resolution can also be beneficial, helping detect smaller anomalies. We additionally include a comparison of inference speed and memory usage across various resolutions in Table~\ref{tab:perf_res}, demonstrating that our model scales effectively with increased resolution. However, we note that beyond a certain point, increasing the resolution offers diminishing returns given the associated increase in computational load.

\begin{table*}[t]
    \small
    \begin{minipage}[t]{.32\linewidth}
    \centering
    \begin{tabular}{lcc}
    \toprule
         & Det.  & Loc. \\
         \midrule
         $256 \times256$ & 93.6 & 87.4 \\
         $512 \times 512$ & 95.4 & 91.4 \\
         \bottomrule
    \end{tabular}
    \caption{Average Anomaly Detection and Localisation results ( AUROC / AUPRO ) for VisA dataset at different resolutions. Increased resolution in the presence of tiny anomalies in VisA leads to an increase in performance.}
    \label{tab:det_res} 
    \end{minipage}
    \hfill
    \begin{minipage}[t]{.65\linewidth}
        \centering
        \begin{tabular}{lccc}
        \toprule
         & $256 \times256$ & $512 \times 512$ & $1024 \times 1024$ \\
             \midrule
        Inference T. [ms] & 9.5 $\pm$ 0.1  & 16.2 $\pm$ 0.3 & 46.0 $\pm$ 0.1 \\
        FPS [img/s] & 105 $\pm$ 0.63 & 61.6 $\pm$ 1.3 & 21.7 $\pm$ 0.0 \\
        Memory [MB] & 413.1 $\pm$ 48.5 & 542.0 $\pm$ 27.3 & 1551.6 $\pm$ 27.1 \\
             \bottomrule
        \end{tabular}
        \caption{Inference time, frames per second (FPS), and memory usage for different input image resolutions. SuperSimpleNet scales well with increased resolution, maintaining fast operation with 21 images per second and reasonable memory usage of 1.5 GB even when processing a megapixel ($1024\times 1024$) image.}
        \label{tab:perf_res}    
    \end{minipage}
\end{table*}

\noindent\textbf{Misclassification analysis.}

The results of misclassification analysis are presented in Table~\ref{tab:missc}. We find that false positives are relatively rare, and typically occur on images that exhibit minor imperfections, which may plausibly be considered defective. Representative examples are shown in Figure~\ref{fig:fail_qual} (top-right quadrant - \textcolor{red}{red square}). The two capsule samples contain small, unclean regions that are difficult to spot without zooming. The one on the left is labelled as anomalous, while the one on the right is not. The right false positive could reasonably be interpreted as an anomaly based on visual similarity to an actual defect case in the right capsule. With a thorough inspection of data, we found that almost all false positives exhibit this property. This highlights a labelling ambiguity, which is often rooted in subjective interpretation by expert annotators, a known challenge also discussed by the author of the dataset in Section 5.3 of \cite{racki_sensum}.

For false negatives, our quantitative findings (summarized in Table~\ref{tab:missc}, top two rows) show that the dominant factor is anomaly size. Most FNs correspond to extremely small defects, typically covering less than 1~\% of the image area. This limitation arises from the limited resolution of the input features. However, we show in the ablation study (shown in Table~\ref{tab:det_res}) that increasing input resolution leads to performance improvements, suggesting that false negatives caused by tiny defects can be mitigated by higher-resolution processing.

\begin{table*}[t]
    \small
    \centering
    \begin{tabular}{llcccccc}
    \toprule
    ~ & Dataset & TP & TN & FP & FN & \makecell{TP mean \\ ano. ratio} & \makecell{FN mean. \\ ano. ratio} \\
    \midrule
    \multirow{2}{*}{\textbf{Ours}}  &  SensumSODF & 432 & 1641 & 41 & 66 & 1.92 & 0.72 \\
        ~ & KSDD2 & 95 & 890 & 4 & 15 & 3.38 & 0.98 \\
    \midrule
    \multirow{2}{*}{Without Synth. Ano.} &SensumSODF & 430 & 1630 & 52 & 68 & 1.98 & 0.38 \\
       ~ & KSDD2 & 94 & 888 & 6 & 16 & 3.38 & 1.11 \\
    \bottomrule
       
    \end{tabular}
    \caption{Misclassification analysis. For each dataset and setup, we provide a count of individual types of classification as well as the mean ratio of image covered by an anomalous region for true positive and false negative predictions. We can observe two things: false negative predictions predominantly come from very small anomalies covering less than 1~\% of the image. The second observation is that synthetic anomalies help reduce misclassification.}
    \label{tab:missc}
\end{table*}

To analyse the effect of synthetic anomalies on misclassification, we compare the number of false positives and false negatives with and without synthetic data in Table~\ref{tab:missc}. The top two rows show results with real and synthetic data, and the bottom two rows show results without synthetic data. This comparison focuses on datasets where real anomalies are present in the training set (e.g., KSDD2 and SensumSODF), since MVTec AD and VisA follow the unsupervised protocol and are trained using only synthetic anomalies. 
The results confirm that the addition of synthetic data helps reduce misclassification. On SensumSODF, the number of false positives decreases by 11 samples (a 21~\% reduction) when synthetic anomalies are used. In case of KSDD2, the false positive count decreases from 6 to only 4 (a 33~\% reduction). This suggests that synthetic anomalies act as a regulariser, exposing the model to a wider variety of anomalies and helping it better distinguish rare but benign patterns from true defects. 
We also observe a modest reduction in the number of false negatives. While this effect is less pronounced, it is expected, as false negatives are primarily caused by very small defects, which are not explicitly addressed by synthetic anomalies.

\section{Conclusion}

We proposed SuperSimpleNet, a novel discriminative anomaly detection model tailored to meet industrial requirements for high performance, speed, and flexibility across diverse supervision scenarios involving labels at varying levels of annotation. As the first unified model to excel in unsupervised, weakly supervised, mixed, and fully supervised settings, SuperSimpleNet uniquely utilises all available training data—including images and different types of annotations—to deliver superior and consistent results.

SuperSimpleNet demonstrates state-of-the-art results across all supervision scenarios. In the supervised setting, the method is evaluated on two well-established benchmarks, SensumSODF and KSDD2, achieving 98.0\% AUROC on SensumSODF and 97.8\% $\text{AP}_{\text{det}}$ on KSDD2. On SensumSODF and KSDD2, SuperSimpleNet surpasses all previous methods by 1.1~\% and 1.6~\%, respectively. Weakly supervised setting is evaluated on the same two datasets, achieving 97.4\% AUROC on SensumSODF and 97.2 $\text{AP}_{\text{det}}$ on KSDD2, outperforming previous methods on SensumSODF and KSDD2 by 5.9~\% and 7.9~\%, respectively.  Similarly, in the mixed supervision setting, it consistently outperformed previous methods under varying levels of labelled data. It also achieves state-of-the-art results in the unsupervised setting on two well-established benchmarks, MVTec AD and VisA, with 98.3\% and 93.6\% AUROC, respectively. It achieves these results with highly efficient performance, achieving an inference time of just 9.5 ms and a throughput of 262 images per second.

The main limitation of our method is the dependence on the pretrained feature extractor. If the extracted features do not hold the necessary information to sufficiently model the normality of the object, it will deteriorate downstream anomaly detection performance. Additionally, although we show that the hyperparameters are robust across a wide variety of models, they may need adjustments for some specific real-world cases. Another limitation arises in the detection of very tiny anomalies, where a higher resolution is required for successful operation.

Our results strongly indicate that practical anomaly detection applications should utilise all available information, including images and labels at different levels of annotation, to achieve optimal performance. The findings also underscore the promise of combining knowledge from unsupervised and supervised approaches to enhance detection capabilities. Furthermore, they highlight the advantages of mixed supervision as an effective trade-off between performance and annotation effort. By unifying diverse supervision paradigms while maintaining strong efficiency and reliability, SuperSimpleNet offers a promising approach for developing models that address the complex needs of real-world industrial applications.

\subsection*{Acknowledgements}

This work was in part supported by ARIS research projects L2-3169 (MV4.0), GC-0001 (AI4Sci), and J2-60055 (MUXAD), research programme P2-0214, and SLING (ARNES, EuroHPC Vega). 

\bibliography{sn-bibliography}
\clearpage

\begin{appendices}

\section{Numeric results of mixed supervision}

In this section, we also provide the numeric results for the experiments with mixed supervision. Results for SensumSODF are in Table~\ref{tab:sensum_mixed_det} and~\ref{tab:sensum_mixed_loc}, and for KSDD2 in Table~\ref{tab:ksdd_mixed_det} and~\ref{tab:ksdd_mixed_loc}.

\begin{table}[!h]
    \centering
    \resizebox{\linewidth}{!}{
        \begin{tabular}{lccccccc}
		\toprule
		 Model & unsup& 0& 0.2& 0.4& 0.6& 0.8& 1\\ \midrule
		SimpleNet & \meanwithstd{71.1}{3.77}& -& -& -& -& -& \meanwithstd{88.4}{1.84}\\
		TriNet & -& 91.5& 92.2& 92.5& 93.9& 95.2& 96.9\\
		\textbf{Ours} & \meanwithstd{89.9}{0.82}& \meanwithstd{97.4}{0.11}& \meanwithstd{97.8}{0.28}& \meanwithstd{97.8}{0.19}& \meanwithstd{97.8}{0.10}& \meanwithstd{97.8}{0.11}& \meanwithstd{98.0}{0.19}\\

        \bottomrule
        \end{tabular}
    }\caption{Results of anomaly detection with mixed supervision on the SensumSODF dataset (AUROC)}
    \label{tab:sensum_mixed_det}
\end{table}

\begin{table}[!h]
    \centering
    \resizebox{\linewidth}{!}{
        \begin{tabular}{lccccccc}
		\toprule
		 Model & unsup& 0& 0.2& 0.4& 0.6& 0.8& 1\\ \midrule
		SimpleNet & \meanwithstd{34.7}{1.02}& -& -& -& -& -& \meanwithstd{89.6}{1.14}\\
		TriNet & -& -& -& -& -& -& -\\
		\textbf{Ours} & \meanwithstd{89.5}{0.31}& \meanwithstd{92.8}{2.12}& \meanwithstd{94.5}{0.65}& \meanwithstd{95.1}{0.52}& \meanwithstd{94.9}{0.30}& \meanwithstd{95.7}{0.49}& \meanwithstd{95.8}{0.28}\\

        \bottomrule
        \end{tabular}
    }\caption{Results of anomaly localisation with mixed supervision on the SensumSODF dataset (AUPRO)}
    \label{tab:sensum_mixed_loc}
\end{table}

\begin{table}[!h]
    \centering
    \resizebox{\linewidth}{!}{
        \begin{tabular}{lcccccc}
		\toprule
		 Model & unsup& 0& 16& 53& 126& 246\\ \midrule
		SimpleNet & \meanwithstd{45.4}{22.29}& -& -& -& -& \meanwithstd{93.5}{1.05}\\
		DSR & 87.2& -& 91.4& 94.6& -& 95.2\\
		SegDecNet & -& 73.3& 83.2& 89.1& 92.4& 95.4\\
		MaMiNet & -& 80.0& 89.7& 92.3& 94.1& 96.2\\
		\textbf{Ours} & \meanwithstd{55.8}{9.48}& \meanwithstd{97.2}{0.48}& \meanwithstd{97.3}{0.35}& \meanwithstd{97.7}{0.63}& \meanwithstd{97.4}{0.46}& \meanwithstd{97.8}{0.18}\\

        \bottomrule
        \end{tabular}
    }\caption{Results of anomaly detection with mixed supervision on KSDD2 ($\text{AP}_{\text{det}}$). }
    \label{tab:ksdd_mixed_det}
\end{table}

\begin{table}[!h]
    \centering
    \resizebox{\linewidth}{!}{
        \begin{tabular}{lcccccc}
		\toprule
		 Model & unsup& 0& 16& 53& 126& 246\\ \midrule
		SimpleNet & \meanwithstd{0.4}{0.05}& -& -& -& -& \meanwithstd{75.9}{2.40}\\
		DSR & 61.4& -& 71.2& 81.6& -& 85.5\\
		SegDecNet & -& 1.0& 45.1& 52.2& -& 67.6\\
		MaMiNet & -& -& -& -& -& -\\
		\textbf{Ours} & \meanwithstd{45.8}{1.81}& \meanwithstd{47.6}{2.46}& \meanwithstd{59.1}{1.50}& \meanwithstd{68.0}{1.37}& \meanwithstd{76.5}{0.62}& \meanwithstd{81.3}{0.64}\\

        \bottomrule
        \end{tabular}
    }\caption{Results of anomaly localisation with mixed supervision on KSDD2 ($\text{AP}_{\text{loc}}$).}
    \label{tab:ksdd_mixed_loc}
\end{table}

\section{Hyperparameter sensitivity analysis}

\begin{figure}[!h]
    \centering
    \includegraphics[width=1\linewidth]{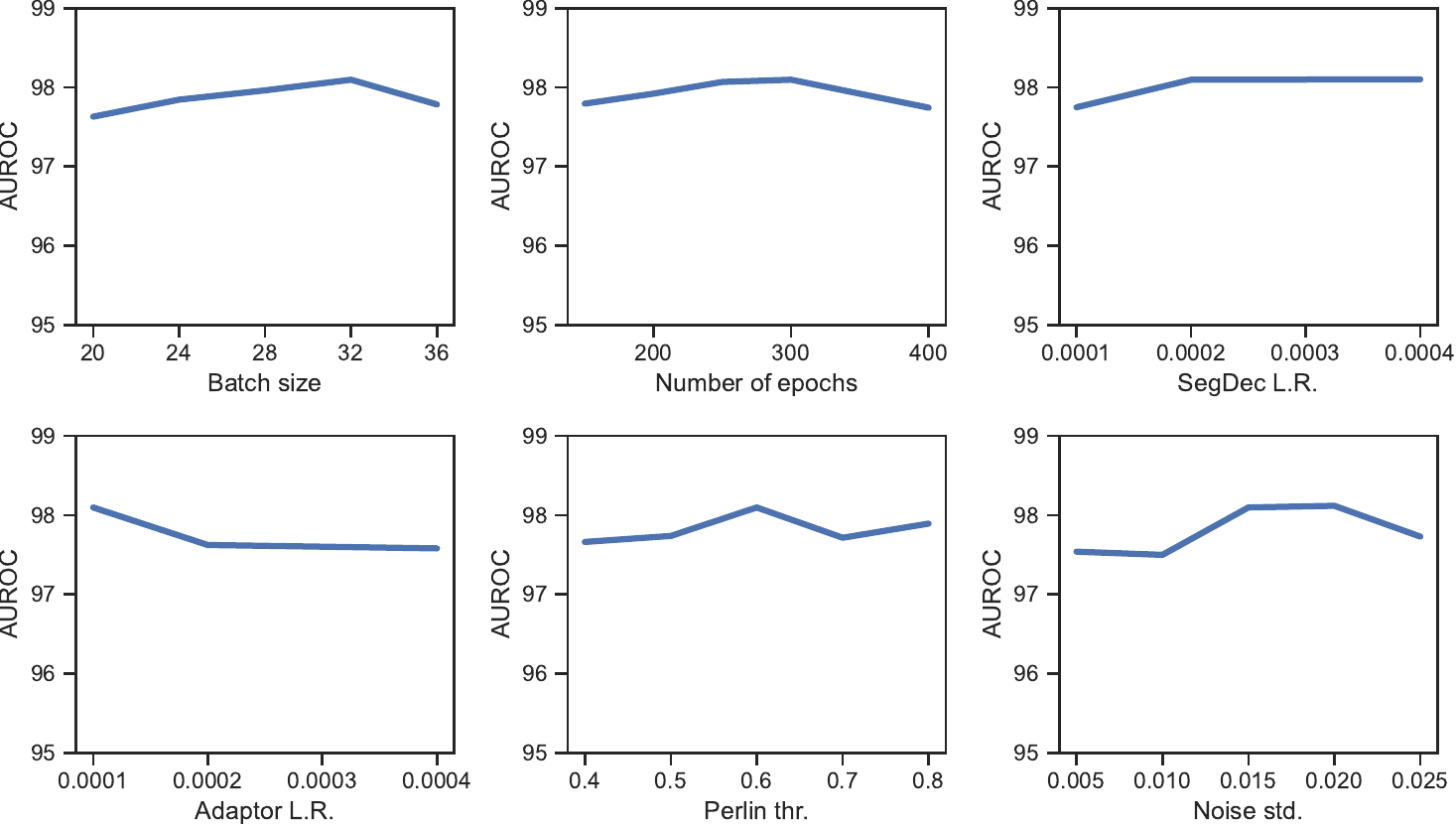}
    \caption{Hyperparameter sensitivity analysis, performed via cross-validation on SensumSODF. The results show that our method is relatively robust to hyperparameter selection, as performance remains stable across a reasonable range around the selected values.}
    \label{fig:hyper}
\end{figure}

To further support future reproducibility and transparency, we present the results of hyperparameter sensitivity analysis in Figure~\ref{fig:hyper}. This analysis was performed by cross-validation on SensumSODF. Our model involves five primary hyperparameters: batch size, learning rate, number of epochs, Perlin mask threshold, and Gaussian noise standard deviation ($\sigma$). 
The results demonstrate that the hyperparameter configuration performs consistently well across the tested ranges, with some values yielding slightly better results.

\end{appendices}

\end{document}